\documentclass[Afour,sagev,times]{sagej}

\usepackage{multirow}
\usepackage{moreverb,url}
\usepackage{bm}
\usepackage[colorlinks,bookmarksopen,bookmarksnumbered]{hyperref}
\usepackage[dvipsnames]{xcolor}
\usepackage{hyperref}
\usepackage[linesnumbered]{algorithm2e}
\SetKwComment{Comment}{// }{}
\RestyleAlgo{ruled}
\usepackage{enumitem}
\usepackage{tabularx}

\definecolor{brand}{RGB}{160,90,109}
\definecolor{emph_color}{RGB}{230, 184, 0}
\definecolor{brand2}{RGB}{107, 148, 134}

\DeclareMathOperator*{\argmax}{arg\,max}

\hypersetup{
  colorlinks=true,
  urlcolor=brand,
  citecolor=brand2
}

\newcommand\BibTeX{{\rmfamily B\kern-.05em \textsc{i\kern-.025em b}\kern-.08em
T\kern-.1667em\lower.7ex\hbox{E}\kern-.125emX}}

\def\volumeyear{2025}

\begin{document}

\runninghead{Dennler, et al.}

\title{Improving through Interaction: \\Searching Behavioral Representation Spaces with CMA-ES-IG}

\author{Nathaniel Dennler\affilnum{1}, Zhonghao Shi\affilnum{2}, Yiran Tao\affilnum{1},\\ Andreea Bobu\affilnum{1}, Stefanos Nikolaidis\affilnum{2} and Maja Matari\'c\affilnum{2}}

\affiliation{\affilnum{1} CSAIL/AeroAstro, Massachusetts Institute of Technology.\\
\affilnum{2} Department of Computer Science, University of Southern California.}

\corrauth{Nathaniel Dennler, dennler@mit.edu}

\begin{abstract}
Robots that interact with humans must adapt to individual users’ preferences to operate effectively in human-centered environments. An intuitive and effective technique to learn non-expert users' preferences is through rankings of robot \textit{behaviors}, e.g., trajectories, gestures, or voices. Existing techniques primarily focus on generating queries that optimize preference learning \textit{outcomes}, such as sample efficiency or final preference estimation accuracy.
However, the focus on outcome overlooks key user expectations in the \textit{process} of providing these rankings, which can negatively impact users' adoption of robotic systems.
This work proposes the Covariance Matrix Adaptation Evolution Strategies with Information Gain (CMA-ES-IG) algorithm. CMA-ES-IG explicitly incorporates user experience considerations into the preference learning process by suggesting perceptually distinct and informative trajectories for users to rank. We demonstrate these benefits through both simulated studies and real-robot experiments. CMA-ES-IG, compared to state-of-the-art alternatives, (1) scales more effectively to higher-dimensional preference spaces, (2) maintains computational tractability for high-dimensional problems, (3) is robust to noisy or inconsistent user feedback, and (4) is preferred by non-expert users in identifying their preferred robot behaviors. 
This project's code is available at \href{https://github.com/interaction-lab/CMA-ES-IG}{github.com/interaction-lab/CMA-ES-IG}
\end{abstract}

\maketitle

\section{Introduction}

\begin{figure*}
    \includegraphics[width=\linewidth]{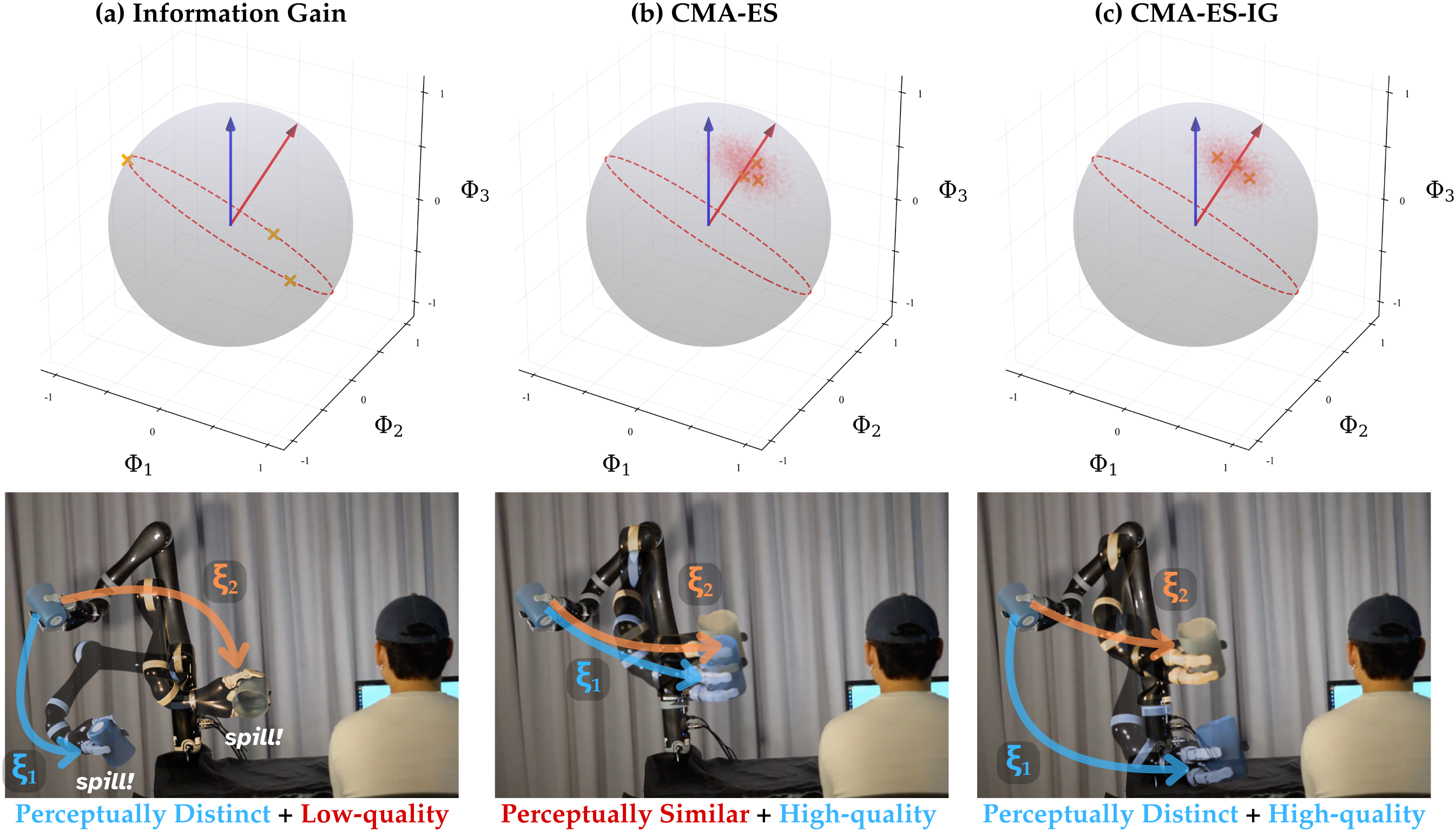}
    \caption{Query Generation Techniques. The spheres on the top row represent a 3-dimensional trajectory feature space, with a blue arrow indicating the user's true preference and the red arrow indicating the current estimated preference. The red circle denotes the equator of the sphere defined by the red preference estimate. The red cloud of points denote the sampling space for CMA-ES. Optimizing for information gain results in trajectories that spaced evenly across the red circle which are easy to distinguish for the user, but do not achieve high rewards (a). Using Covariance matrix adaptation evolution strategies (CMA-ES) results in trajectories with higher rewards, but they are not easy for the user to easily distinguish (b). CMA-ES with information gain (CMA-ES-IG) generates trajectories that are both easy to distinguish for the user and achieve high reward.}
  \label{fig:front-figure}
\end{figure*}

Robots deployed in human-centered environments must adapt their behaviors to align with the preferences and expectations of individual users~\cite{dennler2023design,rossi2017user}. For example, one user may want a robot to hand them an item as quickly as possible, while another may prefer more cautious motions. These preferences can only be learned through interaction with users, however, many users lack the expertise to formally specify desired robot behaviors. In order to align robot behaviors with user preferences, robots should be teachable through interfaces that are both efficient and intuitive.

Human-in-the-loop optimization offers such an interface by enabling users to iteratively guide robot behavior through feedback~\cite{slade2024human}. A particularly effective form of feedback is ranking robot trajectories, which allows non-experts to specify their preferences without requiring programming knowledge~\cite{brown2020better,keselman2023optimizing}. Existing human-in-the-loop optimization approaches generally leverage rankings in two ways: (1) to explicitly learn a model of the user’s reward function~\cite{brown2020better}, or (2) to implicitly infer preferences through black-box optimization~\cite{keselman2023optimizing,lu2022preference}. 
Prior works have shown that these two approaches lead to effective \textit{outcomes} when learning user preferences, but the \textit{process} of teaching the robot may still be unintuitive to users~\cite{Habibian2022Heres}, leading to noisy feedback and reduced learning efficiency. 

When explicitly modeling a user, robots propose trajectories for the user to rank that are \textit{informative} with respect to an underlying feature space. This focus on informativeness can result in trajectories that are perceptually different and thus easy for a user to rank, but the suggested trajectories may be unintuitive for users, who do not directly observe or reason about the underlying feature space. When implicitly modeling a user through black-box optimization approaches, the robot improves over time, but the robot also assumes that users can reliably evaluate rankings over candidate trajectories. However, users often provide noisy or inexact labels, especially when trajectories are perceptually similar. The varied assumptions made in these two types of human-in-the-loop optimization lead to robot teaching processes that are difficult for users to understand.    


To address this gap, we propose an algorithm that combines the strengths of explicit and implicit preference-learning approaches to create an intuitive robot teaching process. Our method, Covariance Matrix Adaptation Evolution Strategy with Information Gain (CMA-ES-IG), generates candidate trajectories that balance informative exploration with perceptual distinguishability for users. In simulation, we examine the scalability of our approach and its application across several preference-based robot tasks. We found that CMA-ES-IG produces candidate trajectories that align more closely with simulated and real user preferences than state-of-the-art baselines. We then evaluated the algorithm’s real-world applicability in experiments where users taught robots to perform both physical and social tasks. In the physical task, participants taught a JACO2 arm how to hand over objects; in the social domain, they taught a Blossom robot~\cite{suguitan2019blossom,o2024design,shi2024build} to create state-expressive gestures. Across both domains, we find that CMA-ES-IG effectively learns user preferences while improving the quality of robot trajectories over time. More broadly, this work highlights the importance of designing optimization objectives that jointly optimize preference-learning accuracy and the user’s experience of the robot teaching process.

\section{Related Work}

In this section, we review prior work on the role of user preferences in physically and socially assistive robotics, and we review existing methods for learning these preferences through human-in-the-loop optimization to situate our work relative to previous approaches.

\subsection{Preferences in Physically Assistive Robots}
Physically assistive robots (PARs) help users with tasks such as collaborative assemblies~\cite{nemlekar2023transfer,gombolay2018fast}, physical rehabilitation~\cite{zhang2017human}, and activities of daily living~\cite{bhattacharjee2020more,dennler2021design}. Although the final goals of these tasks are often objective, differences in user preferences for \textit{how} such goals are accomplished can significantly affect efficiency in human–robot collaboration~\cite{zhang2017human,nikolaidis2015efficient}. For example, prior works have shown that automatic adaptation to user preferences through user clustering can improve efficiency in assembly tasks\cite{nemlekar2023transfer,nikolaidis2015efficient}.

However, other works have found that users also value maintaining control over how a robot assists them in physical tasks. In particular, users can perceive fully autonomous physically assistive systems as reducing their own agency~\cite{bhattacharjee2020more,collier2025sense}. In order to maintain user agency when interacting with these systems, researchers have employed personalization procedures in systems that assist users with eating~\cite{canal2016personalization}, reaching for items~\cite{jeon2020shared}, and object handovers~\cite{perovic2023adaptive}, thereby improving users’ perceptions of the system and its ability to perform tasks. In our real world study, we expand the prior work on robot handovers by explicitly examining how the preference-learning process affects user perception.

\subsection{Preferences in Socially Assistive Robots} Socially assistive robots (SARs) are designed to assist users through primarily social interaction, rather than physical means~\cite{mataric2016socially}. SARs have been used to facilitate behavior change in a wide variety of user populations, including college students~\cite{o2024design,kian2024can}, children with learning differences~\cite{saleh2021robot,shi2023evaluating}, users with limited mobility~\cite{dennler2021personalizing,dennler2023metric}, and the elderly~\cite{ghafurian2021social,zhou2021designing}. Due to the unique and varied needs of these populations, system designers must develop robot behaviors that are specifically tailored to each user population~\cite{clabaugh2019escaping}. 

However, robot system developers cannot realistically design a specific system for each individual user. Therefore, to make robots useful to diverse populations, users must personalize robot behaviors to their own preferences~\cite{gasteiger2023factors}. Previous works have found several benefits of adapting robot behaviors to the user~\cite{rossi2017user}, though these works often require researchers to modify the robot based on knowledge of the participant~\cite{martins2019user,tapus2008user}. Allowing the user to modify the robot’s behaviors themselves, such as through selecting voice~\cite{shi2023evaluating}, personality~\cite{tapus2008user}, task challenge level~\cite{clabaugh2019long}, or gesture~\cite{moro2018learning} can lead to increased acceptance of the system. 

\subsection{Learning User Preferences}
There are many ways for users to provide feedback to learn user preferences~\cite{fitzgerald2022inquire}. In general, robots can learn a user’s internal reward function through \textit{inverse reinforcement learning} by modeling this reward function as a mapping from a low-dimensional representation of behaviors to a scalar reward value. Various techniques can be used to learn this mapping from trajectory to reward, including demonstrations~\cite{abbeel2004apprenticeship}, physical corrections~\cite{bajcsy2017learning}, language~\cite{sharma2022correcting}, trajectory rankings~\cite{brown2020better}, and trajectory comparisons~\cite{sadigh2017active}. Users may need varying levels of expertise with controlling robot systems to effectively use those techniques~\cite{fitzgerald2022inquire}. Our work focuses on using {\it trajectory rankings}, a technique that is accessible to users of any level of expertise.

There are two common approaches for learning user preferences from rankings. One approach is to explicitly model the user’s reward function by estimating a probability distribution over reward parameters~\cite{sadigh2017active,biyik2018batch}. This type of approach uses probabilistic models of user ranking behaviors to update the estimated distribution over reward weights using Bayes' rule. The Bayesian approach considers the full hypothesis space of trajectories at each iteration of querying the user. By explicitly modeling users' reward functions, robot policies can maneuver cars in simulation~\cite{sadigh2017active} and assist users with assembly tasks~\cite{nemlekar2023transfer}. 

The other main approach is to implicitly model the user's reward function using rankings as inputs to derivative-free optimization algorithms. This type of approach directly identifies the trajectory features that the user will rank highly. CMA-ES~\cite{hansen2003reducing} is a performant technique that demonstrates efficiency and tolerance to noise, and has been applied to learning human user preference in robotics domains. CMA-ES has been applied to optimize and identify personalized control laws for exoskeleton assistance that minimizes human energy cost during walking~\cite{zhang2017human}. Recent work also applied CMA-ES to support user preference and goal learning through pairwise selection in social robot navigation~\cite{keselman2023optimizing} and haptic feedback~\cite{lu2022preference}.

Both types of approaches effectively elicit user preferences, but prior works have focused on using these methods in isolation. Additionally, these approaches have primarily focused on optimizing the final performance or accuracy of the robot’s learned preferences, rather than on the user's process of teaching the robot. Importantly, Habibian et al.~\cite{Habibian2022Heres} demonstrated that incorporating users’ perceptions is crucial when robots learn preferences over interpretable, hand-crafted features. In this work, we integrate explicit and implicit user reward models to learn preferences over data-driven feature representations, while explicitly evaluating how users perceive the robot’s teaching process.

\section{Problem Formulation}

\textbf{Preliminaries.} We consider \textit{robot behaviors} as trajectories in a fully-observed deterministic dynamical system.
We denote a behavior as $\xi \in \Xi$, which represents a series of states and actions: $\xi = (s_0,a_0,s_1,a_1,...,s_T,a_T)$ for a finite horizon of $T$ time steps. 
These states and actions are abstractly defined. Traditionally, these sequences refer to robot joint states where the behavior refers to a robot motion, but they can also be interpreted as sequences of images~\cite{liang2025clam} (i.e., videos), sequences of audio frequencies~\cite{yoo2024poe} (i.e., sound), or sequences of tactile feedback~\cite{huang20243d}.

We model a user's preference as a reward function over the space of robot behaviors that maps behaviors to a real value: $R_H : \Xi \mapsto \mathbb{R}$. 
The user's reward function is not directly observable, but it can be inferred through interaction. The goal of preference learning is to estimate a reward function from user interactions, $R_H$, that explains the user's observed behavior.
Higher values of $R_H$ for a particular behavior implies that the behavior is more preferred by the user.

\medskip
\noindent\textbf{Learned Trajectory Representations.} Because the state space of robot behaviors can be very large~\cite{robinson2023robotic,arulkumaran2017deep}, directly learning $R_H$ from state-action sequences is typically intractable. 
To make reward learning more efficient, several works~\cite{bobu2024aligning,ng2000algorithms,abbeel2004apprenticeship} assume that there exists a function $\Phi : \Xi \mapsto \mathbb{R}^d$ that maps from the state-action space to a lower-dimensional \textit{representation space}---a real vector of dimension $d$. This vector is often referred to as the trajectory's \textit{features}.

The specific representation space can take many forms. For example, handcrafted feature spaces are representation spaces that are explicitly defined, such as ``minimum distance to an obstacle" over a trajectory. Representation spaces can also be learned from data through a variety of representation learning techniques such as principal component analysis~\cite{cully2019autonomous,safonova2004synthesizing}, variational auto-encoders~\cite{poddar2024variational,marta2023variquery,zintgraf2021varibad}, or contrastive learning~\cite{bobu2023sirl,dennler2025clea}. In this work, we assume access to the output of the mapping function $\Phi$, regardless of its specific implementation.

\medskip\noindent\textbf{Modeling Users' Reward Functions.} Different users have different preferences for how a robot behaves, corresponding to different reward functions. Following previous work~\cite{sadigh2017active,yang2025trajectory,nemlekar2022towards}, we assume that a user's reward function can be expressed as a linear combination of trajectory features:
\begin{equation}
    R_H(\xi) = \omega^\top\Phi(\xi)
\end{equation}

We aim to estimate the values of $\omega$ for a particular user. Modeling $R_H$ as a linear function is a reasonable assumption because the features, $\Phi$, can be arbitrarily complex~\cite{vapnik2013nature,bobu2024aligning}.

\medskip\noindent\textbf{Modeling Users' Rankings.} We propose to learn $\omega$ from rankings of robot behaviors. A user observes a ranking \textit{query}, $Q$, consisting of $K$ robot behaviors, denoted as $Q = \{\xi_1, \xi_2,...,\xi_K\}$. The user returns an ordered set of these behaviors $\tilde{Q} = \{\xi_{p_1}, \xi_{p_2},...,\xi_{p_K}\}$, where $p_1$ is the index of the most-preferred trajectory, and $p_K$ is the index of the least preferred trajectory. This ranking behavior is modeled as an iterative sampling of $Q$ for the top-preferred trajectory without replacement. Users are assumed to be noisily rational in this sampling process. The probability of selecting a particular trajectory at each iteration is modeled using the Luce-Shepard choice model~\cite{shepard1957stimulus}:
\begin{equation}
    P(\xi_i | Q, \omega) = \frac{e^{R_H(\xi_i)}}{\sum\limits_{\xi_j\in Q}e^{R_H(\xi_j)}}
\end{equation}
Assuming that each choice is independent, we can express the probability of the ranking as:
\begin{equation}\label{eq:luce_shepard}
    P(\tilde{Q} | Q, \omega) = \prod\limits_{i=1}^K P(\xi_{p_i} \mid Q \backslash\{\xi_{p_1}, ...,\xi_{p_{i-1}}\}, \omega)
\end{equation}
This is known as the Plackett-Luce ranking model~\cite{plackett1975analysis,myers2022learning}.

\medskip\noindent\textbf{Updating Preference Estimates.} The probability models assume access to the user's preference to explain the choices a user makes. However, our goal is to estimate the value of $\omega$. To estimate this value, we can maintain a belief distribution over the space of possible $\omega$ values. When we observe a ranking $\tilde{Q}$ from a user, we update the belief by applying Bayes' rule, assuming conditional independence between successive rankings:

\begin{equation}\label{eq:weight_update}
    b_{t+1}(\omega) = \frac{p(\tilde{Q}\mid Q,\omega) b_t(\omega)}{\int p(\tilde{Q} \mid Q, \omega')b_t(\omega')d\omega'}
\end{equation}

In practice, this belief can be represented by a parameterized distribution (e.g., a normal distribution) or by using sample-based techniques (e.g., sequential importance sampling). 

\section{Generating Queries}
Previous work enables a robot to learn a user's preference given a query $Q$ and the user's ranking. The focus of this work is to compute a query $Q$ such that the robot’s suggested behaviors reflect perceptible improvement to the user over time. 

\subsection{Information Gain}

Optimizing for a query's information gain is a technique that leads to bounded regret in the robot's estimate of a user's preference parameters~\cite{biyik2019asking}. This is achieved by solving the following optimization problem:

\begin{equation}\label{eq:infogain}
    Q^* = \argmax\limits_Q H(\tilde{Q} \mid Q, \hat{\omega}) - \mathbb{E}_\omega \left[ H(\tilde{Q} \mid Q, \omega )\right]
\end{equation}

Where $H$ denotes the Shannon information entropy~\cite{shannon1948mathematical} and $\hat{\omega}$ denotes the robot's \textit{maximum a posteriori} (MAP) estimate of $\omega$.

\noindent\textbf{Geometric Interpretation.}  The first term in \autoref{eq:infogain}, $H(\tilde{Q} | Q,\hat{\omega})$, encourages the \textit{robot's} current preference estimate to be maximally uncertain about which ranking $\tilde{Q}$ the user will provide. According to the Luce-Shepard choice model, this occurs when all trajectories in the query achieve the same reward with respect to the current MAP estimate of the user's preference $\hat{\omega}$. For linear reward functions, trajectories receive identical rewards when their feature vectors $\Phi(\xi_i)$ lie on a hyperplane defined by the normal $\hat{\omega}$. This is advantageous to the robot, because having a user rank these trajectories breaks the ``ties" that occur between the trajectories under the robot's current estimate of the user's preference.

The second term in \autoref{eq:infogain}, $- \mathbb{E}_\omega \left[ H(\tilde{Q} \mid Q, \omega )\right]$, seeks to minimize the uncertainty across all possible values of a user's preference $\omega$, thereby reducing ranking noise. This term encourages queries to be easy for the \textit{user} to rank by selecting trajectories that are perceptually distinct. Under the Luce-Shepard choice model, this is achieved when trajectory features $\Phi(\xi_i)$ are maximally distant in the representation space. This objective aims to create queries where the user's ranking is driven by their underlying preference rather than stochasticity caused by indistinguishable options.

This geometric intuition reveals a practical drawback of the pure information gain objective, and we further explicate this intuition in the Appendix. Because the information gain objective prioritizes uncertainty over quality, it leads to undesired behavior near convergence. The objective seeks trajectories orthogonal to the current estimate $\hat{\omega}$, but as the algorithm converges on the user's true preference ($\hat{\omega} \approx \omega^*$), these trajectories are also nearly orthogonal to the user's true preference. When this occurs, the trajectories that compose the query achieve near-zero rewards. Consequently, the user perceives a lack of progress; even as the robot's internal belief becomes more accurate, the observed behaviors fail to demonstrate the high-reward performance the user expects.

\subsection{CMA-ES}
Derivative-free optimization algorithms enable direct search over continuous representation spaces for preferred trajectories~\cite{lu2022preference,keselman2023optimizing}. Among these, the Covaiance Matrix Adaptation Evolution Strategy (CMA-ES) is a widely used and competitive methods for derivative-free optimization in continuous domains of moderate dimensionality~\cite{hansen2010comparing}. CMA-ES is particularly well suited for human-in-the-loop optimization, as this algorithm only requires ordinal rankings of sampled candidate behaviors rather than numerical function values, which are often impractical to obtain from human evaluators expressing preferences.

CMA-ES optimizes objective functions through an adaptive local search process. At each iteration, candidate behaviors are sampled from a multivariate Gaussian distribution, $\mathcal{N}(m, C)$, parameterized by a mean vector $m$ and covariance matrix $C$. The sampled candidates are ranked according to their objective function values. This ranking is then used to update the distribution parameters: the mean is shifted toward regions associated with preferred robot behaviors, while the covariance matrix is adapted to increase exploration in directions that are likely to lead to preferred behaviors. CMA-ES maintains a history of mean values from each iteration, referred to as an \textit{evolution path}, in order to update the covariance matrix while maintaining consistency of rankings across iterations. 

The drawback of directly using CMA-ES for human-in-the-loop optimization of behavioral preferences is that it does not account for human perception of similarity. The process of sampling from a multivariate normal distribution often results in behaviors that have features close to the current mean $\mu$, resulting in behaviors that are \textit{perceptually similar} to human users. This similarity results in increased noise in the ranking process, and limits the effectiveness of CMA-ES.

\subsection{CMA-ES-IG}
We view the information gain objective and CMA-ES optimization as \textit{complementary} query generation techniques. The information gain objective creates sets of robot behaviors that are perceptually distinct and robust to ranking noise, whereas CMA-ES generates sets of robot behaviors that iteratively improve according to a user's preference. We propose CMA-ES-IG to leverage both of these benefits.

The key insight of CMA-ES-IG is that na\"ively sampling from a multivariate normal distribution does not explicitly ensure perceptual discriminability required for human-in-the-loop ranking. To optimize the information gain from each interaction, we employ a quantization-based pruning strategy~\cite{lloyd1982least}. We partition the Gaussian sampled by CMA-ES using K-means clustering and select the cluster centroids to form the query set. This encourages the proposed trajectories to be sufficiently diverse to be easily ranked by a user. We provide our algorithm for generating queries in \autoref{alg:cmaesig}.

\begin{algorithm}[t]
\caption{CMA-ES-IG}\label{alg:cmaesig}
\textbf{Given} a function that generates trajectory features $\Phi$, a number of items to ask the user $K$, a desired number of samples $D$, and a prior belief over user preferences $b_0$\;

\textbf{Initialize} the CMA-ES optimizer with $\mu,C$\; 

\While{user not done}{
$S \gets$ $D$ samples from $\mathcal{N}(\mu, C)$\;
$Q \gets$ cluster centroids obtained by applying KMeans$(S, K)$\;
$\tilde{Q} \gets $ User's ranked responses\; 
$b_{t+1} \propto b_t \cdot p(\tilde{Q} \mid Q, \omega)$\;
Update $\mu,C$ according to CMA-ES update~\cite{hansen2003reducing}\;
}
\end{algorithm}

\section{Experiments}

In our experiments, we seek to answer the following research questions: (1)~Does CMA-ES-IG improve overall performance in simulation across problems with different dimensions compared to baseline methods? (2)~Does CMA-ES-IG consistently improve overall performance across representation spaces created by different representation learning methods? (3) Do the benefits of CMA-ES-IG transfer from simulated users to real users?

These questions are designed to successively relax common assumptions in human-centered robot learning algorithms. RQ1 assumes a simple representation structure and a linear user reward function across different feature space dimensions. RQ2 evaluates robustness across multiple representation structures while still assuming access to a known user reward function.
RQ3 evaluates performance in learned representation spaces without assuming access to the user’s reward function, which may be nonlinear in practice.

\subsection{Evaluation Metrics}\label{sec:evaluation_metrics}
To evaluate interactive optimization algorithms in simulation (RQ1 and RQ2), we leverage two types of metrics: (1) a set of metrics that assess the \textit{accuracy} of the user's estimated preference compared to the ground-truth preference, and (2) a set of metrics that assess \textit{improvement} of the candidate trajectories proposed by the robot over time.

To measure \textbf{accuracy} of an estimated preference, we calculate an \textit{alignment} value, which captures parameter-level agreement, and a \textit{regret} value, which captures reward-level performance metric. Together, these metrics capture complementary aspects of learned reward accuracy~\cite{wilde2023dowe}. \textit{Alignment} is a value from -1 to 1 that is defined as the expected cosine similarity of the estimated user preference and the ground truth user preference~\cite{sadigh2017active}:

\begin{equation}
    Alignment = \mathbb{E}_{\omega} \left[ \frac{\omega\cdot\omega^*}{\|\omega\| \cdot\|\omega^*\|} \right]
\end{equation}

\begin{table*}[t]
\centering
\caption{Marginal means for accuracy metrics in the parameter estimation task. We report the mean ($\pm$ standard error) of the area under the curve (AUC) of alignment and regret. These metrics compare the estimated user preference to the ground-truth simulated user preferences, with arrows denoting better values. Bolded values denote the highest-performing method for each experiment.}
\label{tab:quantitative_results}
\resizebox{\textwidth}{!}{%
\begin{tabular}{l cccc cccc}
\toprule
\multirow{2}{*}{\textbf{Algorithm}} & \multicolumn{4}{c}{\textbf{AUC Alignment ($\uparrow$)}} & \multicolumn{4}{c}{\textbf{AUC Regret ($\downarrow$)}} \\ 
\cmidrule(lr){2-5} \cmidrule(lr){6-9}
& $d=4$ & $d=8$ & $d=16$ & $d=32$ & $d=4$ & $d=8$ & $d=16$ & $d=32$ \\ 
\midrule

CMA-ES-IG & $.882 \pm {\scriptstyle .01}$ & $.810 \pm {\scriptstyle .01}$ & $\mathbf{.718} \pm {\scriptstyle .01}$ & $\bm{.517} \pm {\scriptstyle .01}$ & $.265 \pm {\scriptstyle .05}$ & $.492 \pm {\scriptstyle .04}$ & $\bm{.759} \pm {\scriptstyle .08}$ & $\mathbf{1.453} \pm {\scriptstyle .11}$ \\
CMA-ES & $.908 \pm {\scriptstyle .01}$ & $.847 \pm {\scriptstyle .01}$ & $.692 \pm {\scriptstyle .01}$ & $.457 \pm {\scriptstyle .01}$ & $.157 \pm {\scriptstyle .03}$ & $.413 \pm {\scriptstyle .04}$ & $.995 \pm {\scriptstyle .07}$ & $1.876 \pm {\scriptstyle .08}$ \\
Infogain & $\bm{.928} \pm {\scriptstyle .00}$ & $\bm{.849} \pm {\scriptstyle .01}$ & $.606 \pm {\scriptstyle .01}$ & $.374 \pm {\scriptstyle .02}$ & $\bm{.116} \pm {\scriptstyle .01}$ & $\bm{.332} \pm {\scriptstyle .04}$ & $1.243 \pm {\scriptstyle .10}$ & $2.115 \pm {\scriptstyle .11}$ \\
\bottomrule
\end{tabular}
}
\end{table*}

\begin{figure*}
\includegraphics[width=\linewidth]{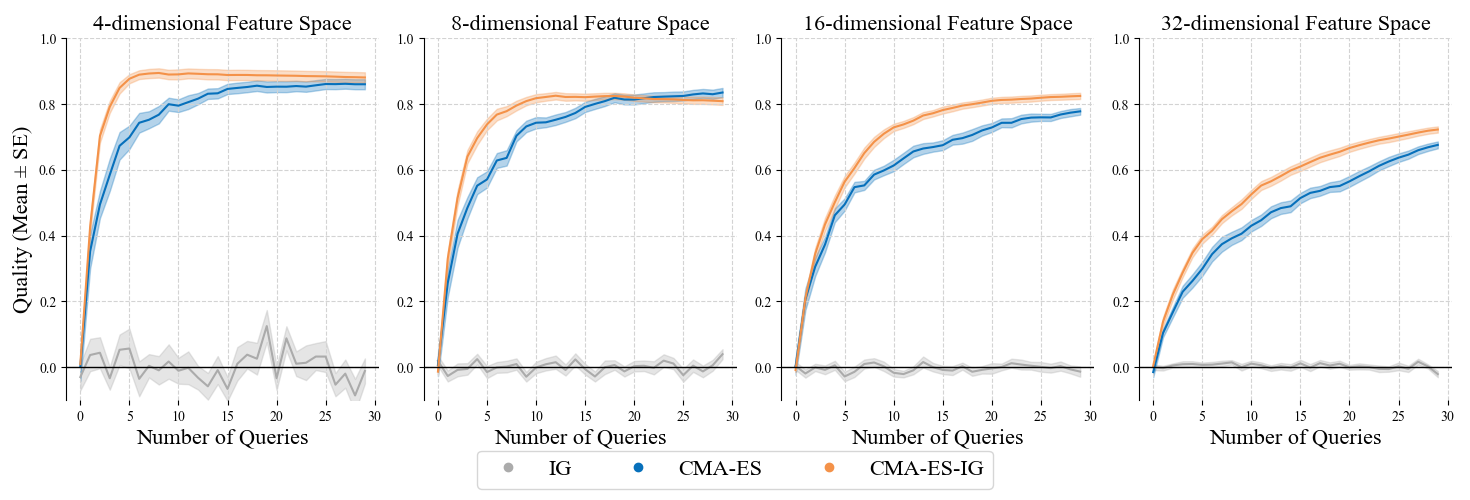}
    \caption{Quality of suggested trajectories over time. Across all dimensions, CMA-ES-IG (orange) generates higher-quality trajectories (i.e., trajectories that receive higher reward on average) for users to rank compared to CMA-ES (blue) and Infogain (gray).}
  \label{fig:parameter_estimation}
\end{figure*}

\textit{Regret} is defined as the difference between the reward achieved by the trajectory that maximizes the user's true preference $\omega^*$ and the reward of the best trajectory selected under the estimated preference. Regret is a positive value and evaluated as:

\begin{equation}
    Regret = R^*(\xi^*) -  R^*(\xi')
\end{equation}

Where $\xi^* = \arg\max_{\xi \in \Xi} \omega^*\cdot\Phi(\xi)$ represents the ground-truth optimal trajectory for a known user preference and $\xi' = \arg\max_{\xi \in \Xi} \hat{\omega}\cdot\Phi(\xi)$ represents the optimal trajectory under an estimated user preference $\hat{\omega}$. $R^*(\cdot)$ represents the simulated user's true reward function.

To measure \textbf{improvement} we propose a metric called \textit{quality} to track the average reward achieved by trajectories that compose the query:

\begin{equation}
    Quality = \frac{1}{\mid Q \mid}\sum_{i=1}^{\mid Q \mid} R(\xi_i | \omega^*)
\end{equation}

Measuring \textit{quality} in this way allows us to track how well-aligned the proposed trajectories are with respect to a user's preference over multiple ranking interactions.

Alignment, regret, and our quality metric change after each interaction with a user. To quantify the evolution of these values across the sequence of interactions, we report the area under the curve for each metric across interaction number, calculated using the composite trapezoidal rule and normalized by the number of interactions in the sequence. This aggregate measure prioritizes algorithms that approach optimal metric values in earlier iterations of the interaction.

\subsection{Scalability Across Feature Dimensions}
Our first research question addresses the problem of searching for samples that align with a user’s preference in a known representation space.
We used the parameter estimation task as described by Fitzgerald et al.~\cite{fitzgerald2022inquire}, which samples a ground truth weight vector, $\omega^*$ from a d-dimensional unit ball. We then generated queries of four trajectories using our three algorithms (Infogain, CMA-ES, and CMA-ES-IG) for a simulated user to rank. Given the ground-truth preference vector of the simulated user, we sampled from the distribution described by Bradley-Terry preference model~\cite{bradley1952rank} to generate rankings. We updated the distribution over the estimated preference vector using Equation \ref{eq:weight_update}. Each experiment simulated 100 users performing 30 ranking iterations for each of the three algorithms. We conducted experiments with feature dimensions $d\in\{4,8,16,32\}$ to evaluate scalability with respect to representation dimensionality.

We formulated the following two hypotheses about the relative performance of these algorithms:
\begin{enumerate}[label=\textbf{H\arabic*.}, ref=H\arabic*]
    \item \label{hyp:rq1_accuracy} CMA-ES-IG will achieve higher AUC for \textit{alignment} and a lower AUC for \textit{regret} scores compared to the baseline algorithms.
    \item \label{hyp:rq1_quality} CMA-ES-IG will demonstrate a higher AUC for \textit{quality} compared to the baseline algorithms.
\end{enumerate}

\noindent\textbf{Results.} To evaluate \textbf{H1}, we conducted pairwise t-tests to compare alignment and regret of CMA-ES-IG to the baseline algorithms. The results of this experiment are presented in Table~\ref{tab:quantitative_results}. We found that Infogain significantly outperformed CMA-ES-IG in low-dimensional spaces ($d\in\{4,8\}$) for both alignment and regret; however, Infogain did not significantly outperform CMA-ES in these settings. For \textit{higher-dimensional} representation spaces ($d\in\{16,32\}$), CMA-ES-IG significantly outperformed Infogain in both alignment and regret. In 32-dimensional representation spaces, CMA-ES-IG also significantly outperformed CMA-ES. \textbf{This result partially supports H1}.

The partial support for \textbf{H1} indicates that CMA-ES-IG more accurately recovers simulated user preferences in higher-dimensional representation spaces, as reflected by improved alignment and lower regret. This pattern is consistent with prior empirical findings that derivative-free optimizers such as CMA-ES are often more competitive in moderate- to high-dimensional optimization problems (roughly 10–500 dimensions), whereas Bayesian Optimization methods tend to perform better in low-dimensional spaces ($d<10$)~\cite{hansen2016cma,santoni2024comparison}. This distinction is particularly relevant given that representation spaces used for learning user preferences are frequently high-dimensional~\cite{bobu2024aligning}. Production-scale preference learning systems routinely employ latent representations with hundreds of dimensions; for example, previous YouTube recommendation systems used 256-dimensional embeddings~\cite{covington2016deep}.

\smallskip To test \textbf{H2}, we conducted t-tests comparing AUC quality of suggested trajectories over time. The graphs of the quality metric after each iteration are shown in Figure~\ref{fig:parameter_estimation}. We found that across all dimensions, CMA-ES-IG achieves a significantly higher AUC for quality compared to Infogain and CMA-ES for all representation space dimensions, with all $p<.01$. \textbf{This result strongly supports H2}.

The support for \textbf{H2} indicates that CMA-ES-IG generates \textbf{\textit{higher-quality}} trajectory queries over sustained interaction with the algorithm. This improvement is consistent with the CMA-ES update mechanism, which iteratively shifts the sampling distribution toward higher-reward regions of the latent space. CMA-ES-IG further outperforms standard CMA-ES by emphasizing perceptually distinct trajectories, reducing ranking noise during user feedback. By ensuring that query trajectories are distinguishable, the algorithm increases the likelihood that user responses accurately reflect their underlying preferences, thereby enabling more reliable updates for the CMA-ES-IG algorithm.

\smallskip Our simulated experiments also found an \textit{additional benefit} of CMA-ES-IG and CMA-ES compared to Infogain. While CMA-ES and CMA-ES-IG incur similar computational costs, implementations of Infogain require solving an optimization problem over candidate queries, which becomes increasingly expensive as dimensionality grows. We found that for high-dimensional problems, CMA-ES-IG and CMA-ES can generate suggested queries using far less computation time than Infogain. For 16-dimensional problems, CMA-ES-IG (4.4 ms) generated queries approximately 500 times faster than Infogain (2287 ms). For 32-dimensional problems, CMA-ES-IG (5.3 ms) generated queries approximately 1000 times faster than Infogain (6256 ms). These results highlight the scalability of CMA-ES-IG to higher-dimensional human-in-the-loop preference learning applications.  

\begin{table}[t]
\centering
\caption{Mean runtime (in milliseconds) for query generation methods. Values are averaged across 100 trials for each representation space dimensions ($d$).}
\label{tab:runtimes}
\resizebox{\columnwidth}{!}{%
\begin{tabular}{l cccc}
\toprule
\textbf{Algorithm} & $d=4$ & $d=8$ & $d=16$ & $d=32$ \\ 
\midrule
CMA-ES-IG & 4.1 ms & 3.9 ms & 4.4 ms & 5.3 ms \\ 
\addlinespace[0.5em]
CMA-ES    & 0.1 ms & 0.1 ms & 0.1 ms & 0.2 ms \\ 
\addlinespace[0.5em]
Infogain  & 418 ms & 805 ms & 2287 ms & 6256 ms \\ 
\bottomrule
\end{tabular}
}
\end{table}

\begin{figure*}
\includegraphics[width=\linewidth]{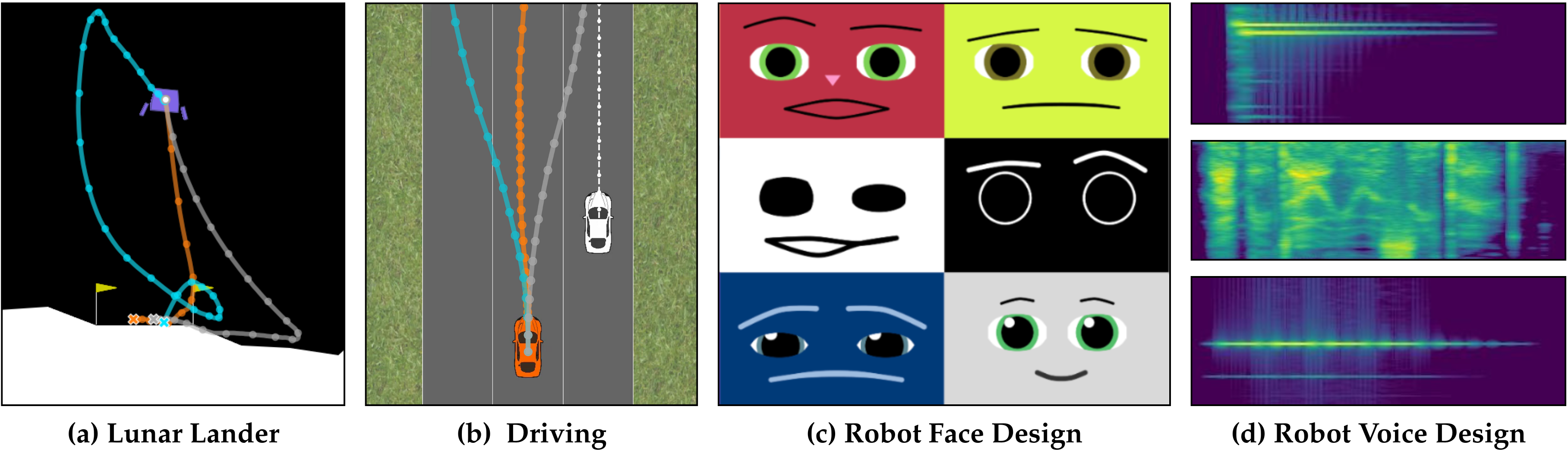}
    \caption{Simulated Domains. We evaluated our algorithms across four simulated domains that represent physical (a,b) and social (c,d) preference-based robot tasks. In Lunar Lander domain (a), preference determines the path of the spaceship takes to land within the flags. In the Driving domain (b), preference determines how the autonomous vehicle merges. In the Face Design domain (c), preference determines the appearance of a screen-based robot face. In the Voice Design domain (d), preference determines the sound of a text-to-speech voice.}
  \label{fig:tasks}
\end{figure*}

\subsection{Performance Across Representation Spaces}
Our previous research question evaluated the scalability of CMA-ES-IG in a representation space that was well-defined and restricted to the unit ball. We aim to evaluate the ability of CMA-ES-IG to generate queries that improve over time in a variety of representation spaces. To explore \textbf{RQ2}, we performed evaluations in four different robotics domains that each use different techniques for defining representation spaces to demonstrate the robustness of CMA-ES-IG. These domains are reflective of tasks common in physical and social robotic applications, and the techniques used to represent trajectories in these domains are reflective of common design choices in robot learning. An overview of these tasks is presented in \autoref{fig:tasks}. 

We found that all choices of representation captured the variance for each domain in four-dimensions. Thus we considered four-dimensional representations for consistency across all simulated robotic tasks. Because these representation spaces may not perfectly align with a unit ball, all candidate trajectories suggested by each algorithm were projected onto a dataset of pre-collected trajectories for each domain. This was achieved by selecting, for each proposed point in representation space, the nearest neighbor (via Euclidean distance) from a dataset of pre-collected trajectories.

\subsubsection{Domain Descriptions}~\\ 
\noindent\textbf{Lunar Lander.} This domain uses the Lunar Lander environment from the Python package ``gymnasium"~\cite{towers2024gymnasium}. The objective of this game is to control a spaceship to land between two flags that denote a goal zone. 
Lunar Lander is often used as an environment to test algorithms in shared autonomy between users and dynamical systems~\cite{reddy2018shared,broad2018learning}. In general shared autonomy settings, users express various preferences for how tasks are completed~\cite{alonso2018system} and deviating from these user preferences can lead to degraded trust~\cite{nikolaidis2017human}, loss of agency~\cite{collier2025sense}, and increased cognitive load~\cite{pan2024effects}, factors that collectively impede the adoption of robots in human-centered environments.

The robot's state in this domain is an 8-dimensional vector that encodes position, velocity, angle, angular velocity, and contact conditions of the robot's legs. The action space consists of four discrete actions: \textit{do nothing}, \textit{fire main engine}, \textit{fire left engine}, and \textit{fire right engine}. A single trajectory lasts for up to 600 time steps, where actions are selected at a frequency of 50Hz. The goal in this domain is to identify a preferred landing trajectory for the Lunar Lander to take to reach the goal.

The feature space in this domain is represented by a randomly initialized neural network that converts a trajectory to a vector in $\mathbb{R}^4$. The collection of trajectories shown to the user was procedurally generated using quality-diversity optimization~\cite{chatzilygeroudis2021quality}. Quality diversity (QD) aims to find a collection of solutions that achieve high objective function (quality) while covering a broad spectrum of behavioral characteristics (diversity). These behavioral characteristics are referred to as a behavior space in the QD literature~\cite{fontaine2020covariance}. In this domain, the traditional Lunar Lander reward maps to the objective function to optimize, and the random neural network representation maps to the axes of the behavior space. Our experiment in the Lunar Lander domain searched for preferences over the QD behavior space.

\medskip\noindent\textbf{Driving.} This domain uses the InterACT driving simulator developed by Sadigh, et al.~\cite{sadigh2016planning}. The goal of this domain is to drive along the highway while avoiding another car on the road. This domain is commonly used to evaluate algorithms for social navigation~\cite{sadigh2016planning} because users have preferences for how robots navigate around them in the physical world~\cite{mavrogiannis2023core}. Robots that navigate in user-preferred ways are more likely to be adopted in human-centered environments~\cite{zhang2021user}.

The state space in this domain is an 8-dimensional vector that incorporates the robot's position, heading, and forward velocity concatenated with the other car's position, heading, and forward velocity. The action space consists of continuous actions that specify angular velocity and linear acceleration. A single trajectory is 200 timesteps at a frequency of 30Hz. The goal of this domain is to identify a user-preferred driving trajectory for driving on a straight road.

In the self-driving domain, we collected a dataset of driving behaviors by randomly sampling robot actions. We trained a variational autoencoder (VAE) on this dataset of robot trajectories to to learn a latent feature representation. Our experiment searched for preferences over the VAE's latent space.

\begin{table*}[t]
\centering
\caption{Mean accuracy metrics across simulated robot tasks. Alignment ($\uparrow$) and Regret ($\downarrow$). All algorithms accurately learned user preferences across all tasks, and CMA-ES-IG demonstrates non-inferiority to the baselines within a 3\% margin across all tasks.}
\label{tab:RQ2_accuracy_results}
\resizebox{\textwidth}{!}{%
\begin{tabular}{l cc cc cc cc}
\toprule
\multirow{2}{*}{\textbf{Algorithm}} & \multicolumn{2}{c}{\textbf{Lunar Lander}} & \multicolumn{2}{c}{\textbf{Driving}} & \multicolumn{2}{c}{\textbf{Robot Face Design}} & \multicolumn{2}{c}{\textbf{Robot Voice Design}} \\ 
\cmidrule(lr){2-3} \cmidrule(lr){4-5} \cmidrule(lr){6-7} \cmidrule(lr){8-9}
& Alignment $\uparrow$ & Regret $\downarrow$ & Alignment $\uparrow$ & Regret $\downarrow$ & Alignment $\uparrow$ & Regret $\downarrow$ & Alignment $\uparrow$ & Regret $\downarrow$ \\ 
\midrule

CMA-ES-IG & $.921 \pm {\scriptstyle .004}$ & $.023 \pm {\scriptstyle .003}$ & $.933 \pm {\scriptstyle .003}$ & $.028 \pm {\scriptstyle .004}$ & $.946 \pm {\scriptstyle .004}$ & $.015 \pm {\scriptstyle .005}$ & $.801 \pm {\scriptstyle .015}$ & $.082 \pm {\scriptstyle .026}$ \\

CMA-ES & $.903 \pm {\scriptstyle .006}$ & $.037 \pm {\scriptstyle .005}$ & $.928 \pm {\scriptstyle .003}$ & $.032 \pm {\scriptstyle .004}$ & $.943 \pm {\scriptstyle .002}$ & $.016 \pm {\scriptstyle .004}$ & $.815 \pm {\scriptstyle .015}$ & $.033 \pm {\scriptstyle .006}$ \\

Infogain & $.933 \pm {\scriptstyle .003}$ & $.019 \pm {\scriptstyle .002}$ & $.948 \pm {\scriptstyle .001}$ & $.016 \pm {\scriptstyle .002}$ & $.960 \pm {\scriptstyle .000}$ & $.005 \pm {\scriptstyle .001}$ & $.852 \pm {\scriptstyle .014}$ & $.017 \pm {\scriptstyle .004}$ \\




\bottomrule
\end{tabular}%
}
\end{table*}

\medskip\noindent\textbf{Robot Face Design.}  This domain uses the robot face animation Python package ``PyLips"~\cite{dennler2024pylips}. The objective of this domain is to select a user-preferred robot face. Screen-based robot faces are a cost-effective way to communicate across embodied channels such as gaze~\cite{admoni2017social,ramnauth2025gaze} and facial expression~\cite{antony2025xpress,elbeleidy2026vizij}. The design of robot faces is an important factor that affects task completion and robot adoption in human-robot collaboration~\cite{kalegina2018characterizing}. 

A state in this domain is a parameterization of the robot's face (e.g., eye separation, mouth size, or color) and the set of action units~\cite{ekman1982methods} that represent the facial expression of the robot at each time step. Actions in this domain refer to changes in the robot's action units. The goal of the simulated human in this domain is to specify their preferred design by selecting the preferred face parameters.

In this domain, we use the hand-crafted features defined in PyLips that control the appearance of the face. To convert this to a four-dimensional space, our experiments randomly sampled four appearance features from PyLips and generated faces that varied within the valid bounds of those features.

\medskip\noindent\textbf{Robot Voice Design.}  This domain uses a multi-speaker text-to-speech model using the Python package ``XTTS"~\cite{casanova2024xtts}. The objective of this domain is to select a user-preferred robot voice. The choice of robot voice can positively affect several outcomes in human-robot interaction~\cite{seaborn2021voice}, including task performance~\cite{ATKINSON2005117}, trust formation~\cite{romeo2025voice}, and cognitive load~\cite{govender2018using}.

The state space in this domain is a set of frequencies and the power at which they are expressed at a given timestep, which are often represented as a vector referred to as the Mel-Frequency Cepstrum Coefficients (MFCC)~\cite{abdul2022mel}. An action in this domain is a change to the MFCCs between time steps. A trajectory is a sequence of MFCCs over time, which is interpreted to users as an audible sound. The goal of this domain is to select a robot's voice that has preferred speech qualities (e.g., pitch, accent, or timbre).

 In the robot voice domain, we leverage the pre-trained speaker embeddings as the behavioral representation space. The speaker embeddings for the XTTS model are a vector in $\mathbb{R}^{512}$ To convert this to a vector in $\mathbb{R}^{4}$ for accurate comparison to the other domains, we use principal component analysis~\cite{greenacre2022principal} to generate a mapping from $\mathbb{R}^{512} \mapsto \mathbb{R}^{4}$ and project all speaker embeddings to this lower-dimensional subspace.

\medskip\noindent\textbf{Hypotheses.} Across these four domains, we formulated the following hypotheses about the relative performance of algorithms when eliciting preferences across a variety of representation spaces:
\begin{enumerate}[label=\textbf{H\arabic*.}, ref=H\arabic*, start=3]
    \item \label{hyp:rq2_non_inferiority} CMA-ES-IG will achieve non-inferior AUC for both \textit{alignment} and \textit{regret} compared to the baseline algorithms by a margin of 0.03.
    \item \label{hyp:rq2_quality} CMA-ES-IG will demonstrate a higher AUC for \textit{quality} compared to the baseline algorithms.
\end{enumerate}

\begin{figure}
    \centering
    \includegraphics[width=\linewidth]{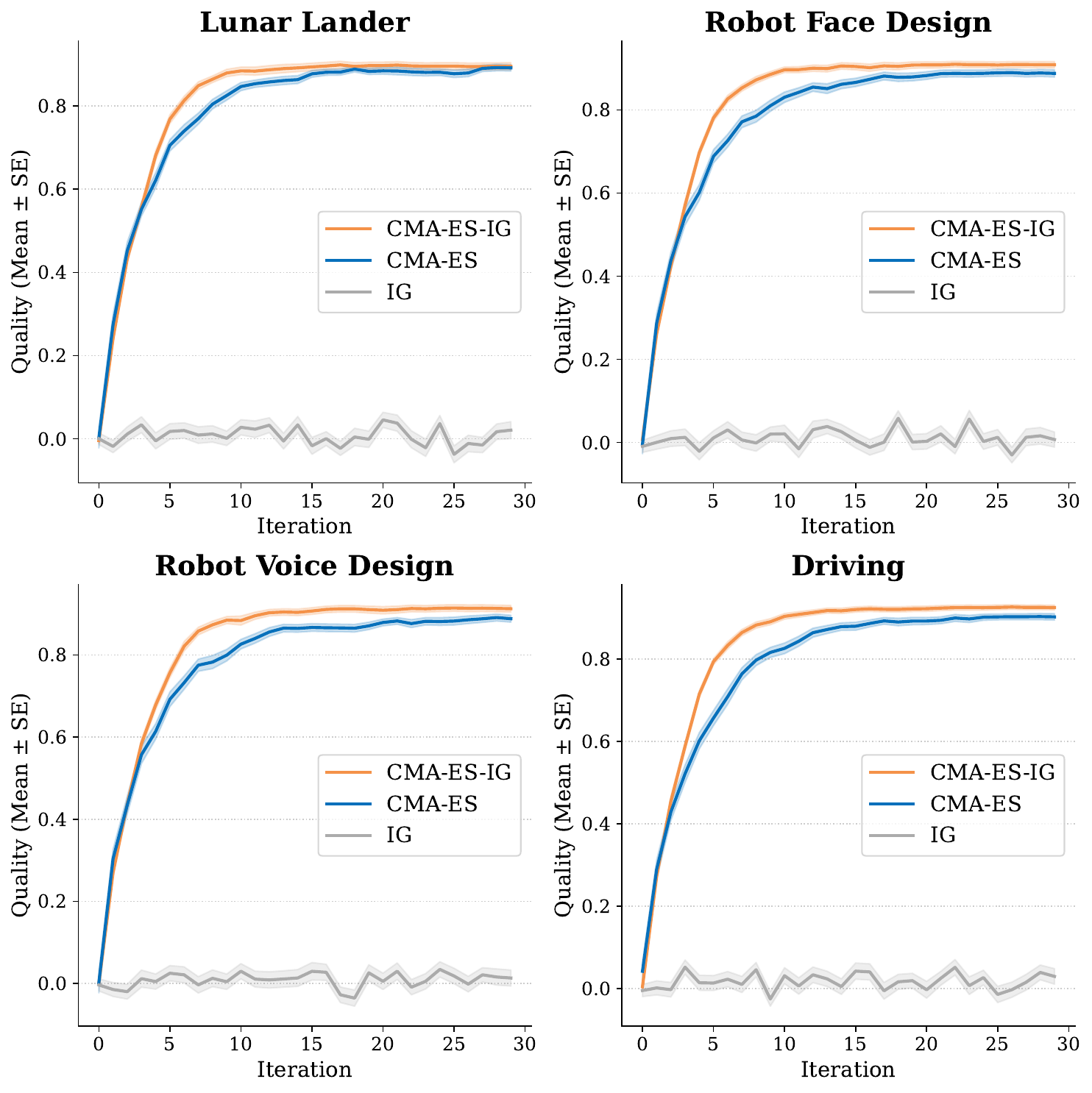}
    \caption{Quality of suggested trajectories over time for simulated robotic environments. For each simulated robot domain, CMA-ES-IG significantly outperforms the baseline algorithms by producing higher-quality trajectory queries in earlier iterations. This result demonstrates that CMA-ES-IG consistently suggests high-quality trajectories under a variety of representation spaces.}
    \label{fig:RQ2_auc_quality}
\end{figure}

\noindent\textbf{Results.} Based on the results from the parameter estimation domain from \textbf{RQ1}, we found that CMA-ES-IG outperforms other baseline algorithms for high-dimensional representation spaces. Because our simulated domains are comprised of low-dimensional representation spaces, \textbf{H3} aims to test non-inferiority in accurately learning user preferences. Non-inferiority testing~\cite{walker2019non} allows us to test whether CMA-ES-IG performs no worse than baseline approaches within a predefined margin. We selected the margin of .03 to represent a small difference in accuracy that is unlikely to be perceptible to a user. The average accuracy metrics for all algorithms are shown in \autoref{tab:RQ2_accuracy_results}. We found that CMA-ES-IG achieves high accuracy, notably closer to Infogain than observed in the parameter estimation task, suggesting that it is robust to choice of representation space. Our non-inferiority tests found that CMA-ES-IG performed no worse than .03 less than the baselines for AUC alignment and AUC regret across all domains with all $p<.001$. \textbf{This result strongly supports H3}.

The results for \textbf{H3} demonstrate that CMA-ES-IG is comparable to CMA-ES and Infogain at learning user preferences across all tested domains. Interestingly, performance across all algorithms was significantly more uniform in these simulated robotics problems than in the parameter estimation task, where Infogain previously exhibited superior alignment and lower regret for idealized 4-dimensional preference learning problems. This shift indicates that CMA-ES-IG and CMA-ES may be more robust to variations and potential distortions in the structure of learned representation spaces. While Infogain’s performance is highly sensitive to the precision of the underlying manifold of representations, the CMA-ES-based approaches appear better equipped to handle the complexities of realistic robotics representations.

\smallskip To test \textbf{H4}, we conducted pairwise t-tests on AUC of quality across the four simulated robotics domains. The results comparing the AUC of quality across robot tasks and algorithms are shown in \autoref{fig:RQ2_auc_quality}. Our statistical tests revealed that CMA-ES-IG achieves significantly higher AUC for quality across all simulated robot tasks compared to CMA-ES and Infogain, all $p<.05$. \textbf{This result provides support for H4}.

The support of \textbf{H4} indicates that CMA-ES-IG generates higher quality trajectory queries across various choices of trajectory representation spaces due to the algorithm's ability to balance information gain with performance. This result extends the finding of \textbf{H2} from structured parameter spaces to the more complex, learned representation spaces that are realistic in robotics problems.

\subsection{Real World Experiments}

The previous research question demonstrated the robustness of CMA-ES-IG to the choice of representation space for various robotic domains. Our next research question seeks to determine if CMA-ES-IG is robust to real user preferences, which may violate the linearity assumption we used in our simulated user models.

\begin{figure}[t]
    \centering
    \includegraphics[width=\linewidth]{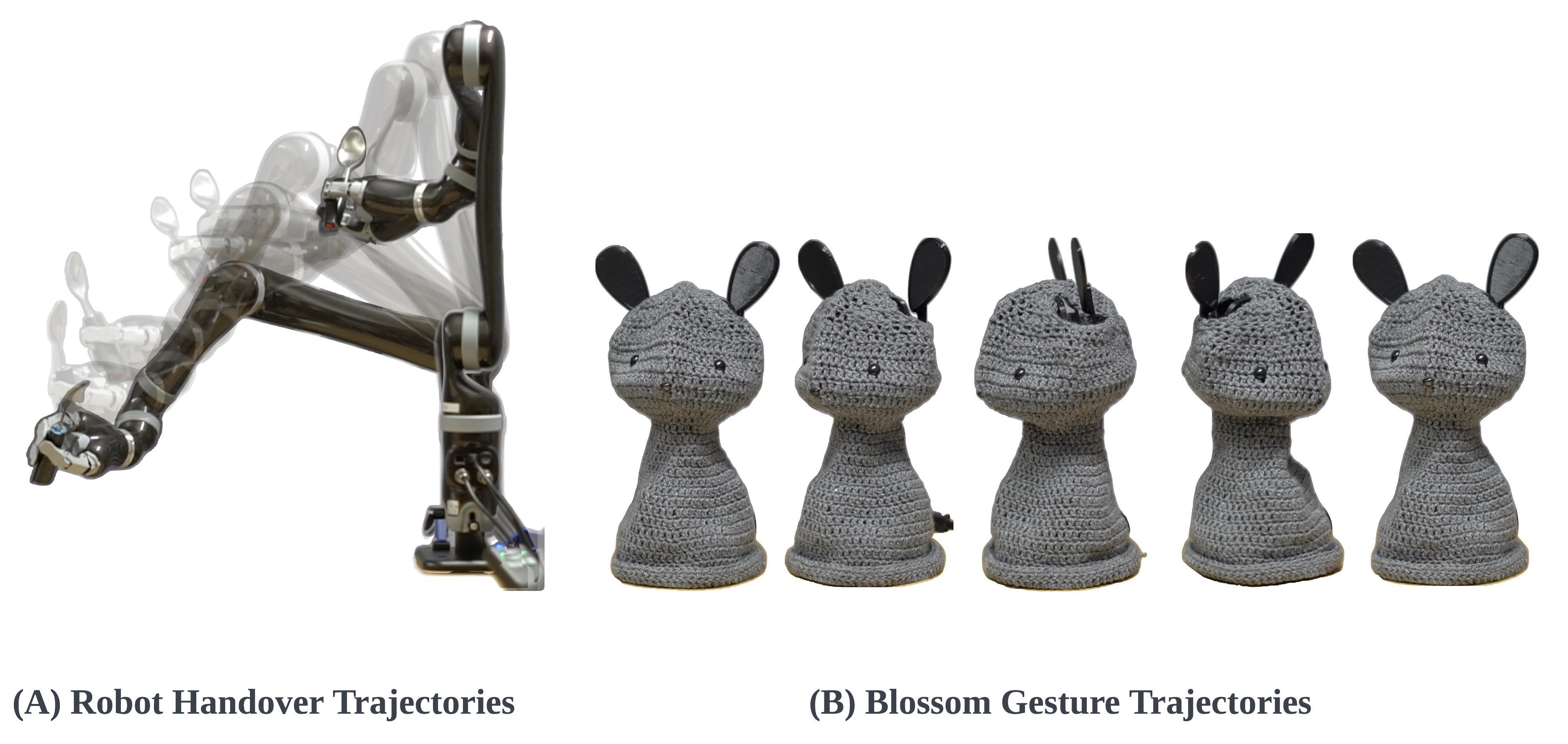}
    \vspace{-2em}
    \caption{The two domains that users taught robots their preferences for the robot's behaviors. In the physical domain (a), users ranked a JACO arm's movement trajectories to hand them a marker, a cup, and a spoon. In the social domain (b), users ranked a Blossom robot's gestures to portray happiness, sadness, and anger. }
    \label{fig:RQ3_tasks}
\end{figure}

To evaluate user perception of different human-in-the-loop optimization methods for preference learning, we conducted a within-subjects user study where users taught robots preferences for physical and social tasks, shown in \autoref{fig:RQ3_tasks}. In the physical task, participants specified preferences for how a JACO robot arm hands them an item. In the social task, participants specified preferences for how a Blossom robot performed state-expressive gestures.

\subsubsection{Task Descriptions}~\\
\noindent\textbf{Robot Handover Task (Physical).} This task performed an assistive robot handover trajectory using a 6DoF Kinova JACO robotic arm. The objective of this domain is to select trajectories that lead to user-preferred handovers that achieve desired trajectory qualities (e.g., appropriate height, desired object orientation, or final approach speed). Robots that hand items to users are often used by users with limited mobility to augment their physical interaction with the world~\cite{grigore2013joint,ortenzi2021object}. This task had three variants where the robot handed the user either a marker, a cup, or a spoon.

The robot's state in this task is a 6-dimensional vector that encodes desired joint positions at each time step. An action in this task refers to a change in desired joint position. A trajectory for this task is a sequence of 50 joint states that were scaled to a fixed time frame of 9 seconds for consistency across trajectories. These trajectories were executed using an impedance controller running at 50Hz, where intermediate joint states were calculated using a b-spline interpolation.

The behavioral representation space was generated from a dataset of 1000 sampled handover trajectories. We used an autoencoder to define nonlinear features across the trajectories in this dataset, which has been shown to be an effective representation space in prior work~\cite{brown2020safe}. Each trajectory corresponded to a 4-dimensional latent vector from this autoencoder to learn user preferences.

\medskip\noindent\textbf{Blossom Gesture Task (Social).} This task was performed using a Blossom robot~\cite{suguitan2019blossom,shi2024build}. Blossom is a table-top socially assistive robot that has been used to support behavior change in users performing cognitive behavioral therapy exercises~\cite{kian2024can,kian2025using}, mindfulness practices~\cite{shi2023evaluating}, peer mediation~\cite{shrestha2025exploring}, and student studying habits~\cite{o2024design}. Gestures are an important embodied communication channel that provides robots the ability to establish a social presence when interacting with users~\cite{huang2013modeling,huang2024gestures,cha2018nonverbal}. This task had three variants that communicated happiness, sadness, and anger.

The robot's state in this task is a 4-dimensional vector that encodes the robot's desired joint position at each time step, which pan, tilt, or translate the robot's head. An action corresponds to a change in the desired joint position. A trajectory is a sequence of joint positions that have been scaled to a fixed time frame of 5 seconds. These trajectories were executed with a joint position controller running at approximately 50Hz, where intermediate joint states were calculated using linear interpolation.

Behavioral representations for this task were generated using a dataset of 1500 sampled gesture trajectories. We used an autoencoder to learn a 6-dimensional latent representation that represented each trajectory. User preferences were elicited using this latent representation space.

\begin{figure*}[t]
    \centering
    \includegraphics[width=\linewidth]{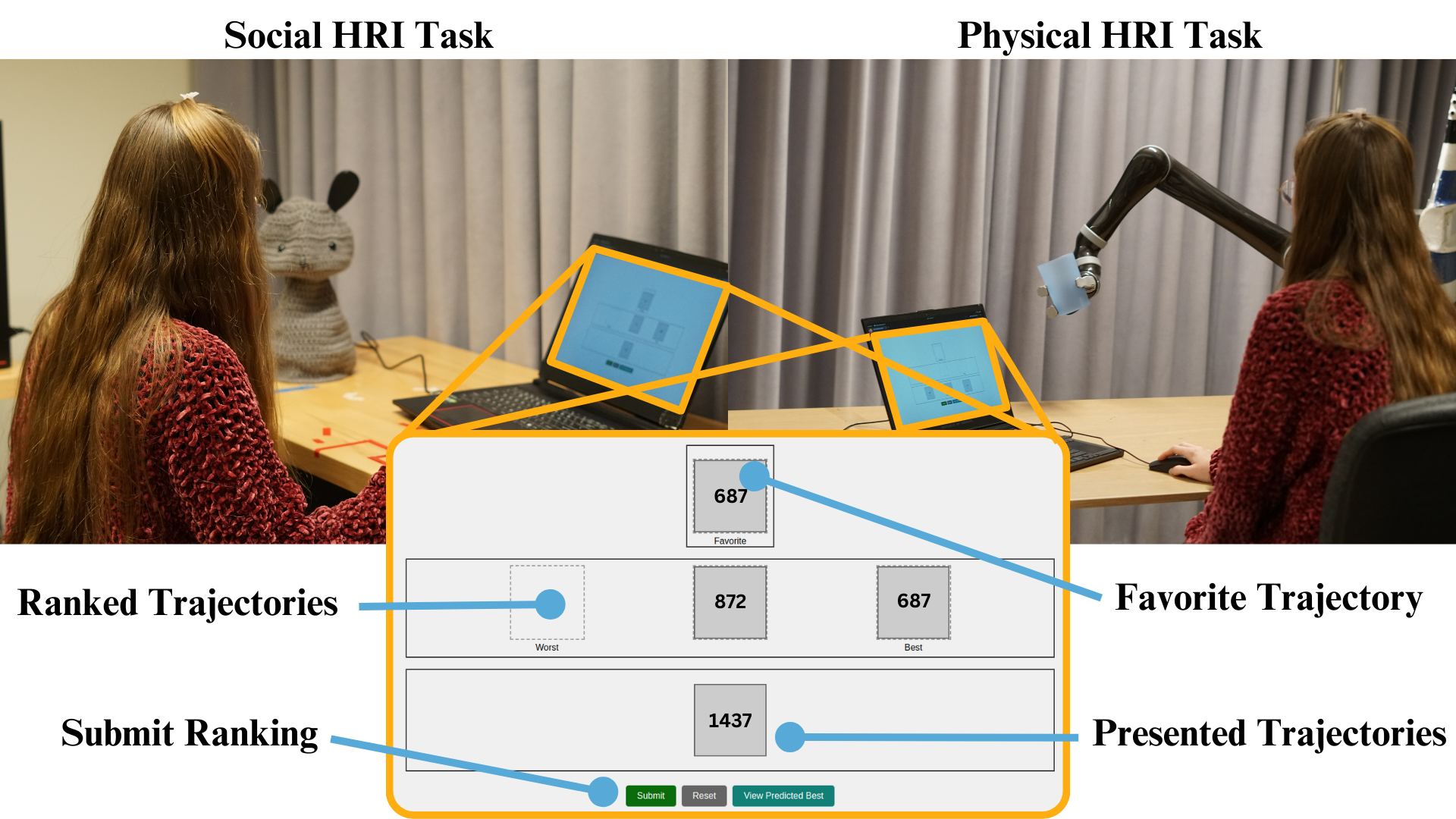}
    \caption{User study interface. Users interacted with the robots through the ranking interface to specify their preferences for how the Blossom robot used gestures to signal different affective states and how the JACO robot arm handed them different items.}
    \label{fig:study_setup}
\end{figure*}

\subsubsection{User Study Details}~\\
\noindent\textbf{Procedure.}
Our within-subject user study tested the three algorithms for generating trajectory queries (Infogain, CMA-ES, and CMA-ES-IG) over three physical (marker handover, cup handover, and spoon handover) and three social tasks (happy gesture, sad gesture, and angry gesture). This study was separated into two blocks that were randomly ordered: a physical task block and a social task block. Within each block, users performed three robot teaching trials, where each trial consisted of a randomly sampled algorithm and task variant. Participants were not informed of the algorithm's name to prevent the name from affecting their ratings. After each trial, participants filled out self-report scales described in the next section. After each block of three trials, participants ranked all three algorithms they had seen relative to each other. Our study procedure was approved by the University of Southern California IRB under \#UP-24-00600.

\medskip\noindent\textbf{Interface.} Each ranking trial that the user engaged in used the same interface, shown in \autoref{fig:study_setup}. The interface sampled three trajectories using one of the three algorithms, and presented these trajectories to the participant. To minimize cognitive load and adhere to human working memory constraints~\cite{cowan2001magical}, each query presented three trajectories to the user ($K=3$). 

The participant used the interface by clicking on boxes to view each candidate trajectory. The participant dragged the boxes to the ranking area, placing the lowest-ranked trajectory in the leftmost box and the highest ranked trajectory in the rightmost box. If a participant found a trajectory particularly aligned with their preferences, they could place it in the reference trajectory box, which was labeled as ``Favorite". This box was included after a pilot study indicated that participants needed a way to refer to trajectories they saw previously. The trajectory in the reference trajectory box was saved for the participant to refer to for the rest of the interaction. Participants could also update their reference trajectory, but only one reference trajectory was active at a time.  Once the participant ranked the three trajectories, they selected the submit button to confirm their preferences. At any time, they could also press the ``View Predicted Best'' button to view the trajectory that maximized their estimated reward.

\medskip \noindent\textbf{Manipulated Variables.} Our study used a 2-by-3 design that manipulated task type (physical and social) and query generation algorithm (CMA-ES-IG, CMA-ES, and Infogain). All participants saw all six conditions.

\medskip\noindent\textbf{Dependent Measures.} As latent user preferences cannot be directly observed, we used validated self-report metrics to evaluate each algorithms' performance. The key contribution to users of CMA-ES-IG is that it generates queries that improve the quality of the suggested trajectories over time (similar to CMA-ES), while additionally being easy to rank (similar to Infogain). We aim to quantify this in our user study through measuring two constructs: \textit{behavioral adaptation} (BA) and \textit{perceived ease of use} (EOU).

BA was measured using the behavioral adaptation scale developed by Lee et al.~\cite{lee2005can}. Perceived behavioral adaptation measures how much the users perceive the robot as changing in response to their inputs. EOU was measured using the ease of use scale created by Venkatesh et al.~\cite{venkatesh2000theoretical}. EOU measures how easily participants are able to interpret and rank the suggest trajectories.
Both scales are psychometrically valid, and measure user perception on a 9-point Likert scale with 0 corresponding to strongly disagree and 8 corresponding to strongly agree. We average across the questions for each of these factors to calculate our evaluation metric.

At the end of each experimental block, users were also asked a forced-ranking question where they ranked the three algorithm they preferred to use. This question measures the user's overall preference for each of the three algorithms we evaluated in the study.

\medskip\noindent\textbf{Hypotheses.} Based on the previous results, we formulated the following hypotheses for our user study:
\begin{enumerate}[label=\textbf{H\arabic*.}, ref=H\arabic*, start=5]
    \item CMA-ES-IG will demonstrate a higher \textit{behavioral adaptation} (BA) score than Infogain.
    \item CMA-ES-IG will demonstrate a higher \textit{ease of use} (EOU) score compared to CMA-ES.
    \item Users will prefer to use CMA-ES-IG compared to the baseline algorithms.
\end{enumerate}

\medskip\noindent\textbf{Recruited Participants.}
 We recruited 14 participants; they were aged 19-32 (Median = 24, SD = 4.5) and comprised 6 women, 5 men, and 3 nonbinary individuals. The cohort identified as Asian ($n=7$), Black ($n=1$), Hispanic ($n=3$), and White ($n=5$); some participants identified as more than one ethnicity. All participants were compensated with a 10 USD Amazon gift card sent to their email. Due to the small sample size, results from the physical and social tasks were aggregated to increase statistical power and ensure a more robust analysis.

\medskip\noindent\textbf{Results.}
To test \textbf{H5}, which posits that CMA-ES-IG demonstrates clearer behavioral adaptation than Infogain, we examine the BA scores for each algorithm, shown in \autoref{tab:RQ3_results}. We first tested internal consistency of the BA scale across users and found that this scale demonstrated very high internal consistency of the scale with a Cronbach's alpha of $\alpha=.97$. We conducted non-parametric Wilcoxon signed-rank test across algorithms. We found that CMA-ES-IG was significantly \textbf{\textit{more adaptive}} than Infogain ($W=5.5$, $p=.009$) with a medium effect size (Hedges' $g=.414$). We additionally found that CMA-ES-IG was rated as significantly more adaptive than the CMA-ES baseline ($W=15$, $p=.033$) with a small to medium effect (Hedges' $g=.377$). \textbf{This result supports H5}.

The confirmation of \textbf{H5} reinforces the insight that users favor CMA-ES-IG because they can observe the robot's behavior actively converging toward their preferences throughout the session. In contrast, while Infogain effectively learns user preferences, its suggestion of trajectories that do not qualitatively improve over time potentially obscures the robot's learning progress from the user. Furthermore, CMA-ES-IG likely outperforms standard CMA-ES because the CMA-ES produces trajectories that are less perceptually distinct. These similarities increase the likelihood of user ranking errors, which inadvertently drives the optimization toward suboptimal regions of the latent space. We further explore the impact of perceptual distinction in \textbf{H6}.

\smallskip To test \textbf{H6}, which posits that CMA-ES-IG is easier to use than CMA-ES, we examined the EOU scores for each algorithm, shown in \autoref{tab:RQ3_results}. We first measured internal consistency of the EOU metric, and found that it demonstrated high internal consistency across users with a Cronbach's alpha of $\alpha=.89$. We again conducted non-parametric Wilcoxon signed-rank test across algorithms. We found that the ratings for CMA-ES-IG were significantly higher than CMA-ES ($W=5.5$, $p=.016$) with a medium effect size (Hedges' $g=.558$).  \textbf{This result supports H6}.

The support of \textbf{H6} indicates that considering perceptual dissimilarity for human-in-the-loop optimization significantly reduces the cognitive burden on the user. Specifically, users found trajectories produced by CMA-ES-IG easier to rank than those from the standard CMA-ES baseline. Interestingly, the lack of a significant difference between CMA-ES-IG and Infogain indicates that perceptual variety is the primary driver of ``rankability"; as long as trajectories are sufficiently distinct, users find the task intuitive, even if the robot's performance does not appear to be improving within that specific query. 

\begin{table}[]
\centering
\caption{Means of Self-reported results. The reported values are aggregated across all physical and social robot tasks. }
\label{tab:RQ3_results}
\resizebox{\columnwidth}{!}{
\begin{tabular}{r|ccc}
\hline
Algorithm & Behavioral Adaptation ($\uparrow$) & Ease of Use($\uparrow$) & Rank ($\uparrow$)\\ 
\hline
CMA-ES-IG & \textbf{5.18} $\pm {\scriptstyle .001}$ & \textbf{5.50} $\pm {\scriptstyle .28}$ & \textbf{1.48} $\pm {\scriptstyle .15}$\\
CMA-ES & 4.69 $\pm {\scriptstyle .42}$ & 4.87 $\pm {\scriptstyle .37}$ & .89 $\pm {\scriptstyle .15}$ \\
Infogain & 4.48 $\pm {\scriptstyle .39}$ & 5.13 $\pm {\scriptstyle .24}$ & .63 $\pm {\scriptstyle .13}$\\
\hline
\end{tabular}
}
\end{table}

\begin{figure}
  \centering
\includegraphics[width=0.48\textwidth]{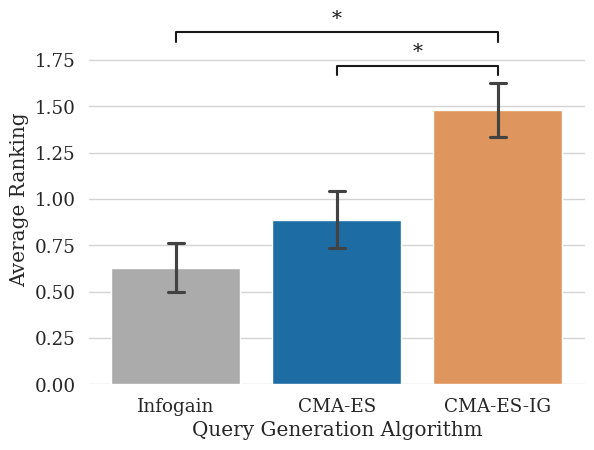}
  \caption{Users' preferences for query generation algorithms. CMA-ES-IG was consistently ranked as the most preferred algorithm for users teaching robots preferences in our user study.}
  \label{fig:rank_results}
\end{figure}

\smallskip To test \textbf{H7}, which posits that users will rank CMA-ES-IG as their most preferred algorithm overall, we examined the rank assigned of each algorithm in the forced-ranking question at the end of each experimental block. A rank of 0 corresponded to the lowest ranked algorithm and a ranking of 2 corresponded to the highest ranked algorithm. The average rank for each algorithm across all users and tasks are shown in \autoref{fig:rank_results}, we found that users ranked CMA-ES-IG as the best algorithm on average ($M=1.48$), followed by CMA-ES ($M=.89$), and Infogain as the least preferred ($M=.63$). Using pairwise Wilcoxon signed-rank tests, we found that CMA-ES-IG was ranked significantly higher than CMA-ES ($W=3.0$, $p=.022$) with a large effect size (Hedges' $g=1.26$). CMA-ES-IG was also ranked higher than Infogain ($W=0.0$, $p=.009$) with a large effect size (Hedges' $g=1.75$). \textbf{This result supports H7}.

Support for \textbf{H7} indicates that the two core principles underlying CMA-ES-IG, iterative improvement and perceptual dissimilarity, create a human-in-the-loop optimization algorithm that is more closely aligned with users' expectations for the teaching process. By satisfying both principles, CMA-ES-IG was significantly preferred by real users over CMA-ES which lacks sufficient perceptual dissimilarity, and Infogain which lacks noticeable improvement across ranking iterations.

\section{Limitations and Future Work}

We evaluated CMA-ES-IG across several simulated and real-world domains; however there are several important limitations of this work to address in future research. A primary constraint of the current study is the reliance on a pre-existing dataset of robot behaviors that successfully perform tasks in diverse ways. In practical applications, curated datasets of high-quality, diverse behaviors are often difficult and costly to collect. To mitigate this dependency, future research can leverage scalable Quality-Diversity (QD) optimization~\cite{tjanaka2022scaling,hedge2023generating}, multi-objective optimization~\cite{abhinav2020multiobjective}, or Large Pre-trained Models~\cite{nguyen2024text} to procedurally generate diverse datasets of robot behaviors. Furthermore, integrating CMA-ES-IG with policy steering frameworks~\cite{wang2025steering,wagenmaker2025steering,du2025dynaguide} could obviate the requirement for offline datasets entirely by optimizing control policies in real-time based on interactive user feedback.

Our current evaluation also relied on task-specific representation spaces. While tailored spaces are effective for isolated interactions, the long-term benefit of assistive robotics lies in their ability to support users across diverse physical and social assistive tasks. Future work should investigate the development of generalized, cross-task representations to facilitate this versatility. Future work in this area may explore learning representations across tasks using multi-task learning frameworks~\cite{yu2024unleashing,taylor2020multitask}, leveraging Large Pretrained Models as semantic priors to transfer preferences across tasks~\cite{barreiros2025careful}, or self-supervised techniques~\cite{poddar2024variational,dennler2025clea} that leverage user-generated data to learn user representations in addition to robot trajectory representations. 

Finally, we acknowledge that our user study was conducted with a convenience sample of university students, which may not fully represent the full demographic of potential end-users for assistive robotics. Future research should prioritize deploying these methods with a wider participant pool, particularly individuals with motor impairments or limited mobility who would most benefit from personalized robotic assistance. Such deployments will uncover critical research questions regarding how preference elicitation can be adapted to alternative input devices used by populations with limited mobility. For instance, future work could investigate how CMA-ES-IG handles the varying noise profiles and bandwidth constraints of gaze-tracking interfaces~\cite{fischer2024scoping,sardinha2024diegetic}, sip-and-puff devices~\cite{javaremi2019interface,nejati2024task}, or natural language voice commands~\cite{padmanabha2024voicepilot,liang2025handproxy}, ensuring that the optimization process remains robust regardless of the input device.

\section{Conclusion}
This work introduced CMA-ES-IG, a human-in-the-loop optimization algorithm to elicit desired robot behaviors. This algorithm addresses two critical desiderata for user-friendly optimization: it generates trajectories that are perceptually distinct for reliable user rankings, and iteratively improve over repeated iterations. Our results demonstrate that CMA-ES-IG accurately recovers user preferences within high-dimensional representation spaces, CMA-ES-IG remains robust across diverse latent structures, and CMA-ES-IG is computationally tractable for high-dimensional representation spaces. Our user study additionally showed that our simulated experiment results robustly transfer to real-world use cases. We hope that this work encourages future algorithms to allow non-expert users to efficiently adapt robot behaviors to their preferences.

\newpage
\bibliographystyle{agsm} 
\bibliography{main}

\newpage
\appendix

\section{Additional Details on Information Gain}
In this section, we provide further formalization for the behaviors that occur when generating queries by optimizing for information gain. We proceed with the assumption that trajectory features are constrained to a d-dimensional unit ball, and the reward function is linear, as in prior work~\cite{sadigh2017active}.

The Infogain objective aims to maximize the mutual information between the preference vector $\omega$ and the observed ranking $\tilde{Q}$:
\begin{equation*}
I(\omega; \tilde{Q} \mid Q) = H(\tilde{Q} \mid Q) - \mathbb{E}_{\omega \sim b_t(\omega)} \left[ H(\tilde{Q} \mid Q, \omega) \right]
\end{equation*}
where $H$ denotes the Shannon entropy. We will examine how each term of this objective behaves for a query  $Q$ consisting of two trajectories $\xi_1$ and $\xi_2$. Our claim is that the first term encourages the query $Q$ to be distributed on a plane orthogonal to $\omega$, and the second term encourages trajectories of $Q$ to maximize pairwise distances in the representation space.

\medskip\noindent\textbf{Orthogonality Claim.}
The first term aims to maximize the robots uncertainty over query $Q$. This is desired because it provides maximum information about $\omega$ by having the user's ranking break ties between trajectories that the robot is uncertain about.  We compute this uncertainty using the Luce-Shepard choice model. For two trajectories, consider the feature difference vector $x = \Phi(\xi_1) - \Phi(\xi_2)$. The probability of a user preferring $\xi_1$ to $\xi_2$ can be rewritten from \autoref{eq:luce_shepard} as:
\begin{equation}
P(\xi_1 \succ \xi_2 \mid Q, \omega) = \frac{1}{1 + \exp(-\omega^\top x)}
\end{equation}
The entropy of a probability distribution over discrete items is maximized when each item has an equal probability. In the case of $Q = \{\xi_1, \xi_2\}$, this implies $P(\xi_1 \succ \xi_2 \mid Q, \omega) = 0.5$ and $P(\xi_2 \succ \xi_1 \mid Q, \omega) = 0.5$. 
Setting the above equation to a value of $0.5$ yields:
\begin{equation}
\frac{1}{1 + \exp(-\hat{\omega}^\top x)} = \frac{1}{2} \implies \exp(-\hat{\omega}^\top x) = 1 \implies \hat{\omega}^\top x = 0
\end{equation}

\textit{Result.} This first term of the information gain objective is maximized when $x$ is \textbf{orthogonal} to the current maximum a posteriori estimate of the user's preference, $\hat{\omega}$.

\smallskip\noindent\textit{Extension to K-item Queries.} In order to achieve a uniform distribution for K-items instead of only two, we must have the following equality across all items in the query:
\begin{equation}
\frac{\exp(\hat{\omega}^\top \Phi(\xi_1))}{\sum \exp(\hat{\omega}^\top \Phi(\xi_j))} = \dots = \frac{\exp(\hat{\omega}^\top \Phi(\xi_K))}{\sum \exp(\hat{\omega}^\top \Phi(\xi_j))}
\end{equation}
This condition is satisfied if and only if all numerators are equal, as denominators are a normalizing constant:
\begin{equation}
\hat{\omega}^\top \Phi(\xi_1) = \hat{\omega}^\top \Phi(\xi_2) = \dots = \hat{\omega}^\top \Phi(\xi_K)
\end{equation}
Rearranging terms, we find that for any pair $(i, j)$:
\begin{equation}
\hat{\omega}^\top (\Phi(\xi_i) - \Phi(\xi_j)) = 0
\end{equation}
\textit{Result.} Any pair of features must be \textbf{orthogonal} to the robots maximum a posteriori estimate of the user's preference $\hat{\omega}$. Conceptually, this requires all trajectories to achieve the same reward.

\medskip\noindent\textbf{Distance Claim.} 
We will again start with the assumption that $Q = \{\xi_1, \xi_2\}$. Consider the value $z=\omega^\top(\Phi(\xi_1) - \Phi(\xi_2))$, which represents the difference in reward of $\xi_1$ and $\xi_2$. We will now denote the probability of selecting $\xi_1$ from the set as:
\begin{equation}
P(\xi_1 \succ \xi_2 \mid Q, \omega) = \frac{1}{1 + \exp(-z)} = \sigma(z)
\end{equation}
where $\sigma$ denotes the sigmoid function. In the two dimensional case, minimizing the entropy results occurs when $\sigma(z) \rightarrow0$ or $\sigma(z) \rightarrow1$, indicating that the user will prefer $\xi_1$ with complete certainty or $\xi_2$ with complete certainty.

Because $\sigma$ is a monotonically increasing function, and the entropy function is concave, $H(\sigma(z))$ is minimized when $|z|$ is large. Since $z$ is defined as $\omega^\top(\Phi(\xi_1) - \Phi(\xi_2))$, this implies that the difference between features $\Phi(\xi_1)$ and $\Phi(\xi_2)$ is maximized.

\textit{Result.} The second term of the information gain objective is minimized when the trajectories are \textbf{maximally distant} in the representation space. Conceptually, this encourages the user's underlying preference signal $|\omega^\top x|$ to overcome the stochastic noise ($\sigma$) of the choice model.

\smallskip\noindent\textit{Extension to K-item Queries}. Consider a query $Q = \{\xi_1, \dots, \xi_K\}$ with feature vectors $\phi_i = \Phi(\xi_i)$. The Shannon entropy of the resulting distribution is minimized when the probability vector approaches a vertex of the $(K-1)$-simplex. Under the Plackett-Luce model, this condition is satisfied when the pairwise probability ratios $P_i / P_j$ are either very large or very small for all pairs $(i, j)$ in the set. This is due to the \textit{Independence of Irrelevant Alternatives} axiom of decision theory, which posits that a users preference between two items will not change when another item is introduced.

Specifically, for a user to be certain in their choice (i.e., have minimum entropy), there must exist one "best" item $\xi_i$ such that for all other items $\xi_j$ in the query, the pairwise reward difference $z_{ij} = \omega^\top(\phi_i - \phi_j)$ is maximized. If even one pair of trajectories remains close in feature space, the user will face a ``tie" between those two options, preventing the total entropy from reaching its minimum.

\textit{Result.} Therefore, the $K$-item case reduces to a simultaneous \textbf{maximization of pairwise distances}. To minimize the second term of the Infogain objective, the robot must select a set of trajectories where the features are maximally dispersed.

\section{Extended Statistical Analyses}

In this section, we present all pairwise comparisons and statistical results for the tests we ran in the experiments section. The effect sizes are reported with the goal that these values can be used by researchers to perform power analyses for future user studies.

\subsection{RQ1 Tables of Comparisons}
RQ1 explored the question \textit{Does CMA-ES-IG improve overall performance in simulation for well-defined representation spaces relative to baseline methods}? We evaluated this question using a continuous parameter estimation task across parameter spaces with different dimensions.

We formulated two hypotheses: (\textbf{H1}) that CMA-ES-IG would achieve higher \textit{AUC alignment} and lower \textit{AUC regret} than baselines, and (\textbf{H2}) that CMA-ES-IG would have a higher \textit{AUC Quality} score compared to baselines. See \autoref{sec:evaluation_metrics} for a description of how these metrics are calculated. For each pairwise comparison, we report the t-statistic, p-value, and Cohen’s $d$. Cohen’s $d$ serves as a measure of effect size and is commonly interpreted as the difference between means in units of standard deviation. In these tables, a negative Cohen’s $d$ indicates that Alg. 1 has a higher mean, and a positive Cohen’s $d$ indicates that Alg. 2 has a higher mean.

The full set of statistical analyses for \textbf{H1} are shown in \autoref{tab:alignment_results} for the \textit{AUC alignment} score and in \autoref{tab:regret_results} for the \textit{AUC regret} scores. These table indicate that for small dimensions (approximately $d<10$), Infogain is a better algorithm for learning user preferences. For higher dimensions (approximately $d > 10$), CMA-ES-IG is the best performing algorithm, achieving higher alignment and lower regret. The separation between algorithms becomes clearer as the dimension increases.  

\newpage
\begin{table}[ht]
\centering
\caption{Alignment results across dimensions. Higher means indicate better policy alignment.}
\label{tab:alignment_results}
\begin{tabularx}{\columnwidth}{l l ccc}
\toprule
\multicolumn{5}{l}{\textbf{Dimension 4}} \\
\multicolumn{5}{l}{Means:  CMA-ES-IG = .88, CMA-ES = .91, Infogain = .93} \\ \midrule
\textbf{Alg. 1} & \textbf{Alg. 2} & \textbf{$t$} & \textbf{$p$}-value & Cohen's \textbf{$d$} \\ 
\midrule
CMA-ES-IG & CMA-ES & -1.60 & $ .118 $ & -.41 \\
CMA-ES-IG & Infogain     & -3.03 & $ .004 $ & -.78 \\
CMA-ES    & Infogain     & -2.42 & $ .020 $ & -.63 \\
\midrule
\multicolumn{5}{l}{\textbf{Dimension 8}} \\
\multicolumn{5}{l}{Means:  CMA-ES-IG = .81, CMA-ES = .85, Infogain = .85} \\ \midrule
\textbf{Alg. 1} & \textbf{Alg. 2} & \textbf{$t$} & \textbf{$p$}-value & Cohen's \textbf{$d$} \\ 
\midrule
CMA-ES-IG & CMA-ES & -2.85 & $ .007 $ & -.74 \\
CMA-ES-IG & Infogain     & -3.16 & $ .003 $ & -.82 \\
CMA-ES    & Infogain     & -.12  & $ .908 $ & -.03 \\
\midrule
\multicolumn{5}{l}{\textbf{Dimension 16}} \\
\multicolumn{5}{l}{Means:  CMA-ES-IG = .72, CMA-ES = .69, Infogain = .61} \\ \midrule
\textbf{Alg. 1} & \textbf{Alg. 2} & \textbf{$t$} & \textbf{$p$}-value & Cohen's \textbf{$d$} \\ 
\midrule
CMA-ES-IG & CMA-ES & 1.82  & $ .075 $ & .47 \\
CMA-ES-IG & Infogain     & 6.72  & $ <.001 $ & 1.73 \\
CMA-ES    & Infogain     & 4.85  & $ <.001 $ & 1.25 \\
\midrule
\multicolumn{5}{l}{\textbf{Dimension 32}} \\
\multicolumn{5}{l}{Means:  CMA-ES-IG = .52, CMA-ES = .46, Infogain = .37} \\ \midrule
\textbf{Alg. 1} & \textbf{Alg. 2} & \textbf{$t$} & \textbf{$p$}-value & Cohen's \textbf{$d$} \\ 
\midrule
CMA-ES-IG & CMA-ES & 3.59  & $ <.001 $ & .93 \\
CMA-ES-IG & Infogain     & 6.72  & $ <.001 $ & 1.73 \\
CMA-ES    & Infogain     & 3.82  & $ <.001 $ & .99 \\
\bottomrule
\end{tabularx}
\end{table}

\begin{table}[ht]
\centering
\caption{Regret results across dimensions. Lower means indicate better performance.}
\label{tab:regret_results}
\begin{tabularx}{\columnwidth}{l l ccc}
\toprule
\multicolumn{5}{l}{\textbf{Dimension 4}} \\
\multicolumn{5}{l}{Means:  CMA-ES-IG = .27, CMA-ES = .16, Infogain = .12} \\ \midrule
\textbf{Alg. 1} & \textbf{Alg. 2} & \textbf{$t$} & \textbf{$p$}-value & Cohen's \textbf{$d$} \\ 
\midrule
CMA-ES-IG & CMA-ES & 1.92  & $ .061 $ & .50 \\
CMA-ES-IG & Infogain     & 2.94  & $ .005 $ & .76 \\
CMA-ES    & Infogain     & 1.37  & $ .178 $ & .35 \\
\midrule
\multicolumn{5}{l}{\textbf{Dimension 8}} \\
\multicolumn{5}{l}{Means:  CMA-ES-IG = .49, CMA-ES = .41, Infogain = .33} \\ \midrule
\textbf{Alg. 1} & \textbf{Alg. 2} & \textbf{$t$} & \textbf{$p$}-value & Cohen's \textbf{$d$} \\ 
\midrule
CMA-ES-IG & CMA-ES & 1.33  & $ .191 $ & .34 \\
CMA-ES-IG & Infogain     & 2.69  & $ .010 $ & .70 \\
CMA-ES    & Infogain     & 1.36  & $ .181 $ & .35 \\
\midrule
\multicolumn{5}{l}{\textbf{Dimension 16}} \\
\multicolumn{5}{l}{Means:  CMA-ES-IG = .76, CMA-ES = .99, Infogain = 1.24} \\ \midrule
\textbf{Alg. 1} & \textbf{Alg. 2} & \textbf{$t$} & \textbf{$p$}-value & Cohen's \textbf{$d$} \\ 
\midrule
CMA-ES-IG & CMA-ES & -2.21 & $ .032 $ & -.57 \\
CMA-ES-IG & Infogain     & -3.92 & $ <.001 $ & -1.01 \\
CMA-ES    & Infogain     & -2.03 & $ .049 $ & -.52 \\
\midrule
\multicolumn{5}{l}{\textbf{Dimension 32}} \\
\multicolumn{5}{l}{Means:  CMA-ES-IG = 1.45, CMA-ES = 1.88, Infogain = 2.12} \\ \midrule
\textbf{Alg. 1} & \textbf{Alg. 2} & \textbf{$t$} & \textbf{$p$}-value & Cohen's \textbf{$d$} \\ 
\midrule
CMA-ES-IG & CMA-ES & -3.14 & $ .003 $ & -.81 \\
CMA-ES-IG & Infogain     & -4.29 & $ <.001 $ & -1.11 \\
CMA-ES    & Infogain     & -1.70 & $ .096 $ & -.44 \\
\bottomrule
\end{tabularx}
\end{table}

\newpage
The full set of statistical analyses for \textbf{H2} are shown in \autoref{tab:per_query_results} for the \textit{AUC quality} score. This table indicates that CMA-ES-IG outperforms baseline algorithms for all dimensions in the parameter estimation task. CMA-ES-IG is able to achieve this performance because it incorporates the CMA-ES update to improve over time, and generates trajectories that minimize ranking noise. In contrast, standard CMA-ES does not explicitly incorporate methods for minimizing ranking noise, and Infogain does not explicitly improve trajectory quality over time.

\begin{table}[ht]
\centering
\caption{Quality results across dimensions. Higher means indicate better performance.}
\label{tab:per_query_results}
\begin{tabularx}{\columnwidth}{l l ccc}
\toprule
\multicolumn{5}{l}{\textbf{Dimension 4}} \\
\multicolumn{5}{l}{Means:  CMA-ES-IG = .82, CMA-ES = .74, Infogain = .00} \\ \midrule
\textbf{Alg. 1} & \textbf{Alg. 2} & \textbf{$t$} & \textbf{$p$}-value & Cohen's \textbf{$d$} \\ 
\midrule
CMA-ES-IG & CMA-ES & 3.35  & $ .002 $ & .86 \\
CMA-ES-IG & Infogain     & 53.21 & $ <.001 $ & 13.74 \\
CMA-ES    & Infogain     & 39.00 & $ <.001 $ & 10.07 \\
\midrule
\multicolumn{5}{l}{\textbf{Dimension 8}} \\
\multicolumn{5}{l}{Means:  CMA-ES-IG = .73, CMA-ES = .69, Infogain = .00} \\ \midrule
\textbf{Alg. 1} & \textbf{Alg. 2} & \textbf{$t$} & \textbf{$p$}-value & Cohen's \textbf{$d$} \\ 
\midrule
CMA-ES-IG & CMA-ES & 3.09  & $ .004 $ & .80 \\
CMA-ES-IG & Infogain     & 70.81 & $ <.001 $ & 18.28 \\
CMA-ES    & Infogain     & 57.46 & $ <.001 $ & 14.84 \\
\midrule
\multicolumn{5}{l}{\textbf{Dimension 16}} \\
\multicolumn{5}{l}{Means:  CMA-ES-IG = .67, CMA-ES = .60, Infogain = .00} \\ \midrule
\textbf{Alg. 1} & \textbf{Alg. 2} & \textbf{$t$} & \textbf{$p$}-value & Cohen's \textbf{$d$} \\ 
\midrule
CMA-ES-IG & CMA-ES & 5.40  & $ <.001 $ & 1.39 \\
CMA-ES-IG & Infogain     & 77.16 & $ <.001 $ & 19.92 \\
CMA-ES    & Infogain     & 56.57 & $ <.001 $ & 14.61 \\
\midrule
\multicolumn{5}{l}{\textbf{Dimension 32}} \\
\multicolumn{5}{l}{Means:  CMA-ES-IG = .53, CMA-ES = .45, Infogain = .00} \\ \midrule
\textbf{Alg. 1} & \textbf{Alg. 2} & \textbf{$t$} & \textbf{$p$}-value & Cohen's \textbf{$d$} \\ 
\midrule
CMA-ES-IG & CMA-ES & 5.04  & $ <.001 $ & 1.30 \\
CMA-ES-IG & Infogain     & 55.34 & $ <.001 $ & 14.29 \\
CMA-ES    & Infogain     & 36.19 & $ <.001 $ & 9.35 \\
\bottomrule
\end{tabularx}
\end{table}

\newpage ~\\
\newpage
\subsection{RQ2 Table of Comparisons}
RQ2 explored the question \textit{Does CMA-ES-IG consistently improve performance across representation learning methods?} Our experiment tested this by evaluating each algorithms performance across four simulated robotics tasks that each used different representation spaces. Due to the simplicity of these tasks, we found that 4-dimensional representation spaces captured all meaningful variation.

We formulated two hypotheses: (\textbf{H3}) that CMA-ES-IG would do \textit{no worse} than the baselines by a margin of .03 for \textit{AUC alignment} and \textit{AUC regret}, and (\textbf{H4}) that CMA-ES-IG would achieve a 
higher \textit{AUC quality} than the baselines. We tested non-inferiority because we previously saw that Infogain outperformed CMA-ES-based approaches for small dimensions.

The full set of statistical analyses for \textbf{H3} are shown in \autoref{tab:alignment_results_tasks} and \autoref{tab:regret_results_tasks}. We found that across all simulated tasks, the three algorithms performed nearly equally.

The full set of statistical analyses for \textbf{H4} are shown in \autoref{tab:quality_results_tasks}. We found that across all simulated tasks, CMA-ES-IG resulted in higher quality trajectories.

\begin{table}[ht]
\centering
\caption{Alignment results across tasks. Higher means indicate better policy alignment.}
\label{tab:alignment_results_tasks}
\begin{tabularx}{\columnwidth}{l l ccc}
\toprule
\multicolumn{5}{l}{\textbf{Lunar Lander}} \\
\multicolumn{5}{l}{Means:  CMA-ES-IG = .92, CMA-ES = .90, Infogain = .93} \\ \midrule
\textbf{Alg. 1} & \textbf{Alg. 2} & \textbf{Diff.} & \textbf{$p$}-value & Cohen's \textbf{$d$} \\ 
\midrule
CMA-ES-IG & CMA-ES & .018 & $ <.001 $ & .37 \\
CMA-ES-IG & Infogain     & -.012 & $ <.001 $ & -.35 \\
\midrule
\multicolumn{5}{l}{\textbf{Driving}} \\
\multicolumn{5}{l}{Means:  CMA-ES-IG = .93, CMA-ES = .93, Infogain = .95} \\ \midrule
\textbf{Alg. 1} & \textbf{Alg. 2} & \textbf{Diff.} & \textbf{$p$}-value & Cohen's \textbf{$d$} \\ 
\midrule
CMA-ES-IG & CMA-ES & .005 & $ <.001 $ & .18 \\
CMA-ES-IG & Infogain     & -.016 & $ <.001 $ & -.74 \\
\midrule
\multicolumn{5}{l}{\textbf{Robot Face Design}} \\
\multicolumn{5}{l}{Means:  CMA-ES-IG = .95, CMA-ES = .94, Infogain = .96} \\ \midrule
\textbf{Alg. 1} & \textbf{Alg. 2} & \textbf{Diff.} & \textbf{$p$}-value & Cohen's \textbf{$d$} \\ 
\midrule
CMA-ES-IG & CMA-ES & -.003 & $ <.001 $ & .198 \\
CMA-ES-IG & Infogain     & -.0135 & $ <.001 $ & -1.36 \\
\midrule
\multicolumn{5}{l}{\textbf{Robot Voice Design}} \\
\multicolumn{5}{l}{Means:  CMA-ES-IG = .93, CMA-ES = .92, Infogain = .94} \\ \midrule
\textbf{Alg. 1} & \textbf{Alg. 2} & \textbf{Diff.} & \textbf{$p$}-value & Cohen's \textbf{$d$} \\ 
\midrule
CMA-ES-IG & CMA-ES & .012 & $ <.001 $ & .31 \\
CMA-ES-IG & Infogain     & -.015 & $ <.001 $ & -.63 \\
\bottomrule
\end{tabularx}
\end{table}

\begin{table}[ht]
\centering
\caption{Regret results across tasks. Lower means indicate better performance.}
\label{tab:regret_results_tasks}
\begin{tabularx}{\columnwidth}{l l ccc}
\toprule
\multicolumn{5}{l}{\textbf{Lunar Lander}} \\
\multicolumn{5}{l}{Means:  CMA-ES-IG = .02, CMA-ES = .04, Infogain = .02} \\ \midrule
\textbf{Alg. 1} & \textbf{Alg. 2} & \textbf{Diff.} & \textbf{$p$}-value & Cohen's \textbf{$d$} \\ 
\midrule
CMA-ES-IG & CMA-ES & -.014 & $ .002 $ & -.35 \\
CMA-ES-IG & Infogain     & .004 & $ <.001 $ & .16 \\
\midrule
\multicolumn{5}{l}{\textbf{Driving}} \\
\multicolumn{5}{l}{Means:  CMA-ES-IG = .03, CMA-ES = .03, Infogain = .02} \\ \midrule
\textbf{Alg. 1} & \textbf{Alg. 2} & \textbf{Diff.} & \textbf{$p$}-value & Cohen's \textbf{$d$} \\ 
\midrule
CMA-ES-IG & CMA-ES & -.004 & $ <.001 $ & -.10 \\
CMA-ES-IG & Infogain     & .012 & $ <.001 $ & .37 \\
\midrule
\multicolumn{5}{l}{\textbf{Robot Face Design}} \\
\multicolumn{5}{l}{Means:  CMA-ES-IG = .01, CMA-ES = .02, Infogain = .00} \\ \midrule
\textbf{Alg. 1} & \textbf{Alg. 2} & \textbf{Diff.} & \textbf{$p$}-value & Cohen's \textbf{$d$} \\ 
\midrule
CMA-ES-IG & CMA-ES & -.002 & $ <.001 $ & -.04 \\
CMA-ES-IG & Infogain     & .010 & $ <.001 $ & .28 \\
\midrule
\multicolumn{5}{l}{\textbf{Robot Voice Design}} \\
\multicolumn{5}{l}{Means:  CMA-ES-IG = .03, CMA-ES = .03, Infogain = .01} \\ \midrule
\textbf{Alg. 1} & \textbf{Alg. 2} & \textbf{Diff.} & \textbf{$p$}-value & Cohen's \textbf{$d$} \\ 
\midrule
CMA-ES-IG & CMA-ES & -.002 & $ <.001 $ & -.08 \\
CMA-ES-IG & Infogain     & .014 & $ <.001 $ & .55 \\
\bottomrule
\end{tabularx}
\end{table}

\begin{table}[ht]
\centering
\caption{Quality results across tasks. Higher means indicate better performance.}
\label{tab:quality_results_tasks}
\begin{tabularx}{\columnwidth}{l l ccc}
\toprule
\multicolumn{5}{l}{\textbf{Lunar Lander}} \\
\multicolumn{5}{l}{Means:  CMA-ES-IG = .63, CMA-ES = .55, Infogain = .01} \\ \midrule
\textbf{Alg. 1} & \textbf{Alg. 2} & \textbf{$t$} & \textbf{$p$}-value & Cohen's \textbf{$d$} \\ 
\midrule
CMA-ES-IG & CMA-ES & 4.42  & $ <.001 $ & .62 \\
CMA-ES-IG & Infogain     & 50.54 & $ <.001 $ & 7.15 \\
CMA-ES    & Infogain     & 39.86 & $ <.001 $ & 5.64 \\
\midrule
\multicolumn{5}{l}{\textbf{Driving}} \\
\multicolumn{5}{l}{Means:  CMA-ES-IG = .66, CMA-ES = .58, Infogain = .01} \\ \midrule
\textbf{Alg. 1} & \textbf{Alg. 2} & \textbf{$t$} & \textbf{$p$}-value & Cohen's \textbf{$d$} \\ 
\midrule
CMA-ES-IG & CMA-ES & 4.15  & $ <.001 $ & .59 \\
CMA-ES-IG & Infogain     & 46.16 & $ <.001 $ & 6.53 \\
CMA-ES    & Infogain     & 42.56 & $ <.001 $ & 6.02 \\
\midrule
\multicolumn{5}{l}{\textbf{Robot Face Design}} \\
\multicolumn{5}{l}{Means:  CMA-ES-IG = .47, CMA-ES = .38, Infogain = .00} \\ \midrule
\textbf{Alg. 1} & \textbf{Alg. 2} & \textbf{$t$} & \textbf{$p$}-value & Cohen's \textbf{$d$} \\ 
\midrule
CMA-ES-IG & CMA-ES & 2.37  & $ .019 $ & .33 \\
CMA-ES-IG & Infogain     & 20.47 & $ <.001 $ & 2.90 \\
CMA-ES    & Infogain     & 12.80 & $ <.001 $ & 1.81 \\
\midrule
\multicolumn{5}{l}{\textbf{Robot Voice Design}} \\
\multicolumn{5}{l}{Means:  CMA-ES-IG = .63, CMA-ES = .54, Infogain = .01} \\ \midrule
\textbf{Alg. 1} & \textbf{Alg. 2} & \textbf{$t$} & \textbf{$p$}-value & Cohen's \textbf{$d$} \\ 
\midrule
CMA-ES-IG & CMA-ES & 4.49  & $ <.001 $ & .64 \\
CMA-ES-IG & Infogain     & 42.51 & $ <.001 $ & 6.01 \\
CMA-ES    & Infogain     & 39.14 & $ <.001 $ & 5.54 \\
\bottomrule
\end{tabularx}
\end{table}

\newpage
\subsection{RQ3 Table of Comparisons}

RQ3 explored the question \textit{Do the benefits of CMA-ES-IG transfer form simulated users to real users}? We evaluated this question through a user study where users specified their preferences for robots performing physical handovers and social gestures.

Because we cannot measure a user's true preference like we can in simulation, we elected to use self-report measurements to assess this research question. We looked at two constructs measured through Likert scales: Perceived Ease of Use (EOU) and Perceived Behavioral Adaptation (BA). We show the specific items that comprised these scales in \autoref{tab:constructs}. We additionally asked users to rank the three algorithms in order of their overall preference for using them.

We formed three hypotheses that tested the intuition we formulated throughout the design of CMA-ES-IG: (\textbf{H5}) that CMA-ES-IG will have a higher BA score than Infogain, (\textbf{H6}) that CMA-ES-IG will have a higher EOU score than CMA-ES, and (\textbf{H7}) that useres will rank CMA-ES-IG as their most preferred algorithm to use.

We show the statistical results for these in \autoref{tab:user_study_results}. We found that CMA-ES-IG was perceived as \textbf{\textit{more adaptive}} than Infogain, CMA-ES-IG was \textbf{\textit{easier to use}} than CMA-ES, and users \textbf{\textit{preferred}} CMA-ES-IG to all other algorithms.

\begin{table}[ht]
\centering
\caption{Likert scale items for our two metrics. Items are rated on a 7-point scale (1: Strongly Disagree, 7: Strongly Agree).}
\label{tab:constructs}
\begin{tabularx}{\columnwidth}{l X}
\toprule
\textbf{ID} & \textbf{Item Content} \\
\midrule

\multicolumn{2}{@{}X@{}}{%
\textbf{EOU} --- Perceived Ease of Use~\cite{venkatesh2000theoretical}
\hfill $\alpha=.89$
} \\
\cmidrule(lr){1-2}
1 & Teaching the robot is clear and understandable. \\
2 & Teaching the robot does not require a lot of mental effort. \\
3 & I find the robot easy to teach. \\
4 & I find it easy to get the robot to do what I want it to do. \\

\midrule
\multicolumn{2}{@{}X@{}}{%
\textbf{BA} --- Perceived Behavioral Adaptation~\cite{lee2005can}
\hfill $\alpha=.97$
} \\
\cmidrule(lr){1-2}
1 & The robot has developed its skills over time because of my interaction with it. \\
2 & The robot's behavior has changed over time because of my interaction with it. \\
3 & The robot has become more competent over time because of my interaction with it. \\
4 & The robot's intelligence has developed over time because of my interaction with it. \\

\bottomrule
\end{tabularx}
\end{table}

\begin{table}[ht]
\centering
\caption{User study subjective metrics and rankings. Higher means indicate better performance/preference.}
\label{tab:user_study_results}
\begin{tabularx}{\columnwidth}{l l ccc}
\toprule
\multicolumn{5}{l}{\textbf{Behavioral Alignment (BA)}} \\
\multicolumn{5}{l}{Means:  CMA-ES-IG = 5.18, CMA-ES = 4.69, Infogain = 4.48} \\ \midrule
\textbf{Alg. 1} & \textbf{Alg. 2} & \textbf{$W$} & \textbf{$p$}-value & Hedges' \textbf{$g$} \\ 
\midrule
CMA-ES-IG & CMA-ES  & 15.0 & $ .032 $ & .38 \\
CMA-ES-IG & Infogain & 5.5  & $ .006 $ & .41 \\
CMA-ES    & Infogain & 41.5 & $ .797 $ & .04 \\
\midrule
\multicolumn{5}{l}{\textbf{Ease of Use (EOU)}} \\
\multicolumn{5}{l}{Means:  CMA-ES-IG = 5.50, CMA-ES = 4.87, Infogain = 5.13} \\ \midrule
\textbf{Alg. 1} & \textbf{Alg. 2} & \textbf{$W$} & \textbf{$p$}-value & Hedges' \textbf{$g$} \\ 
\midrule
CMA-ES-IG & CMA-ES  & 5.5  & $ .011 $ & .56 \\
CMA-ES-IG & Infogain & 23.0 & $ .223 $ & .29 \\
CMA-ES    & Infogain & 28.5 & $ .434 $ & -.29 \\
\midrule
\multicolumn{5}{l}{\textbf{Rank}} \\
\multicolumn{5}{l}{Means:  CMA-ES-IG = 1.48, CMA-ES = .89, Infogain = .63} \\ \midrule
\textbf{Alg. 1} & \textbf{Alg. 2} & \textbf{$W$} & \textbf{$p$}-value & Hedges' \textbf{$g$} \\ 
\midrule
CMA-ES-IG & CMA-ES  & 3.0  & $ .016 $ & 1.26 \\
CMA-ES-IG & Infogain & .0   & $ .004 $ & 1.75 \\
CMA-ES    & Infogain & 15.5 & $ .438 $ & .42 \\
\bottomrule
\end{tabularx}
\end{table}

\begin{figure*}[t]
    \centering
    \includegraphics[width=\linewidth]{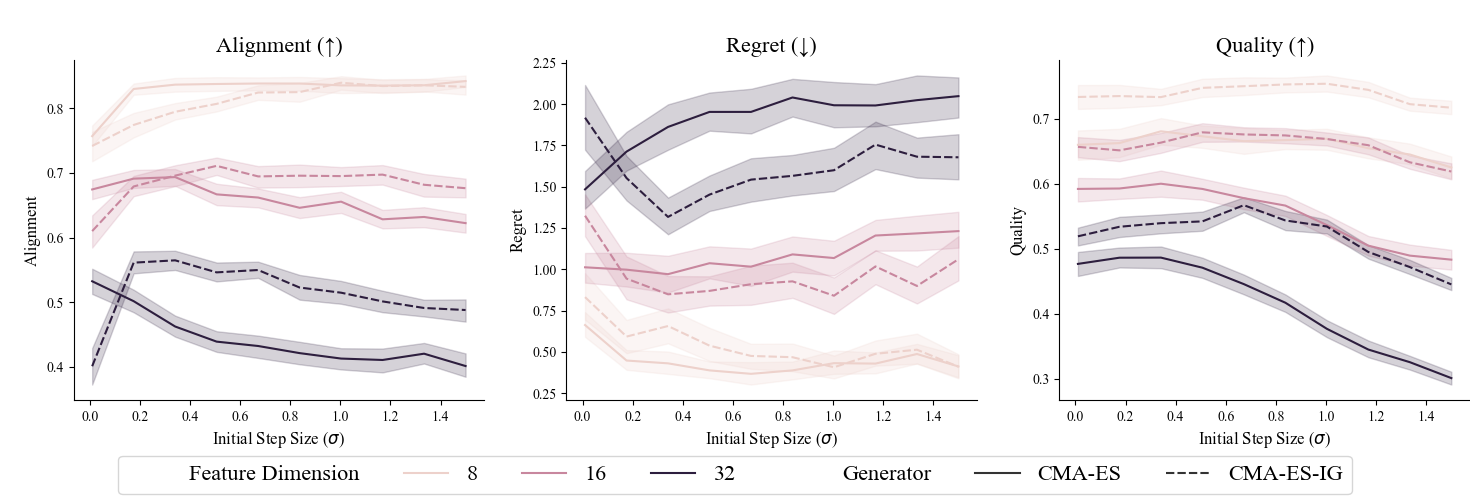}
    \caption{Initial step-size sensitivity. We show the AUC of the labeled metric (corresponding to the values in \autoref{tab:quantitative_results} and \autoref{fig:parameter_estimation}), across 10 different settings of $\sigma$ ranging from 0.01 to 1.5.  We find that the choice of hyperparameter does not largely influence any of the three metrics of interest.  Across most hyperparameter values, CMA-ES-IG outperforms CMA-ES on Alignment and Regret metrics. Across all hyperparameter values, CMA-ES-IG outperforms CMA-ES on the Quality metric.}
    \label{fig:hyperparam_sensitivity}
\end{figure*}

\newpage
\section{Hyperparameter Sensitivity Analysis}
In this paper we explored using CMA-ES and CMA-ES-IG as algorithms for generating queries. We used the default step size parameter of $\sigma=0.5$ which is recommended when there is little information about the structure of the problem space. This ensures that our results are not tuned specifically for our simulation environment, making it more informative of our user study.

We performed an experiment to test varying values of $\sigma$ from 0.01 to 1.5, shown in \autoref{fig:hyperparam_sensitivity}. This experiment used the parameter estimation task from RQ1, and evaluated three metrics: \textit{AUC alignment},\textit{ AUC regret}, and \textit{AUC quality}. 

We found that CMA-ES-IG tended to perform better for higher values of $\sigma$, whereas CMA-ES performed better for lower values of $\sigma$. When selecting optimal values for $\sigma$, CMA-ES-IG outperformed CMA-ES for higher-dimensional feature spaces for \textit{AUC Alignment} and \textit{AUC Regret}. CMA-ES-IG outperformed CMA-ES in \textit{AUC Quality} for nearly all values of $\sigma$.

These insensitivity results are likely due to the CMA-ES automatic step-size adaptation mechanism. This result highlights the strength of CMA-ES-IG to be applicable to a variety of problems without requiring intensive tuning processes.


\end{document}